\documentclass[12pt,preprint,review]{elsarticle}
\usepackage{fullpage}
\journal{Pattern Recognition}
\usepackage[utf8]{inputenc}

\usepackage{algorithm}
\usepackage{algpseudocode}
\usepackage{underoverlap}
\usepackage{multirow}
\usepackage{color}
\usepackage{cancel}
\usepackage{soul}
\usepackage{subcaption}

\usepackage{amsmath, amsfonts}
 
\DeclareMathOperator*{\argmax}{arg\,max}

\biboptions{sort&compress}

\title{Noisy multi-label semi-supervised dimensionality reduction \\
}

\author[UiT,ML]{Karl Øyvind Mikalsen\corref{cor1} }
\address[UiT]{Dept. of Mathematics and Statistics, UiT The Arctic University of Norway, Tromsø, Norway}
\address[ML]{UiT Machine Learning Group}

\cortext[cor1]{Corresponding author at: Department of Mathematics and Statistics, Faculty of Science and Technology,
UiT – The  Arctic University of Norway,
N-9037 Tromsø, Norway}

\ead{karl.o.mikalsen@uit.no}

\author[ML,spain]{Cristina Soguero-Ruiz}
\address[spain]{Dept. of Signal Theory and Comm., Telematics and Computing, Universidad Rey Juan Carlos, Fuenlabrada, Spain}

\author[MLIFT,ML]{Filippo Maria Bianchi}
\address[MLIFT]{Dept. of Physics and Technology, UiT, Tromsø, Norway}

\author[MLIFT,ML]{Robert Jenssen}

\begin{document}

\begin{frontmatter}

\begin{abstract}
Noisy labeled data represent a rich source of information that often are easily accessible and cheap to obtain, but label noise might also have many negative consequences if not accounted for. How to fully utilize noisy labels has been studied extensively within the framework of standard supervised machine learning over a period of several decades. However, very little research has been conducted on solving the challenge posed by noisy labels in non-standard settings. This includes situations where only a fraction of the samples are labeled (semi-supervised) and each high-dimensional sample is associated with multiple labels.
In this work,  we present a novel semi-supervised and multi-label dimensionality reduction method that effectively utilizes information from both noisy multi-labels and unlabeled data. 
With the proposed \emph{Noisy multi-label semi-supervised dimensionality reduction (NMLSDR)} method, the noisy multi-labels are denoised and unlabeled data are labeled simultaneously via a specially designed label propagation algorithm. NMLSDR then learns a projection matrix for reducing the dimensionality by maximizing the dependence between the enlarged and denoised multi-label space and the features in the projected space.
Extensive experiments on synthetic data, benchmark datasets, as well as a real-world case study, demonstrate the effectiveness of the proposed algorithm and show that it outperforms state-of-the-art multi-label feature extraction algorithms.
\end{abstract}


\end{frontmatter}

\section{Introduction}

Supervised machine learning crucially relies on the accuracy of the \textit{observed labels} associated with the training samples~\cite{Nettleton2010, 6685834, Zhu2004, natarajan2013learning, peche, ASLAM1996189, xiao2015learning, lachenbruch1966discriminant, BI20101622, Angluin1988}. 
Observed labels may be corrupted and, therefore, they do not necessarily coincide with the true class of the samples.
Such inaccurate labels are also referred to as \textit{noisy}~\cite{6685834,liu2016classification, natarajan2013learning}. Label noise can occur for various reasons in real-world data, e.g. because of imperfect evidence, insufficient information, label-subjectivity or fatigue on the part of the labeler.
In other cases, noisy labels may result from the use of frameworks such as anchor learning~\cite{Halpernocw011, MIKALSEN2017105} or silver standard learning~\cite{agarwal2016learning}, which have received interest for instance in healthcare analytics~\cite{CALLAHAN2017279, doi:10.1146/annurev-biodatasci-080917-013315}. A review of various sources of label noise can be found in~\cite{6685834}. 
 
In standard supervised machine learning settings, the challenge posed by noisy labels has been studied extensively. For example, many noise-tolerant versions of well-known classifiers have been proposed, including 
discriminant analysis~\cite{lachenbruch1966discriminant,  Lawrence:2001:EKF:645530.655665}, 
logistic regression~\cite{BOOTKRAJANG20143641},
the k-nearest neighbor classifier~\cite{wilson2000reduction}, 
boosting algorithms~\cite{long2010random, mcdonald2003empirical},
perceptrons~\cite{Bylander:1994:LLT:180139.181176, crammer2009adaptive},
 support  vector machines~\cite{biggio2011support},
deep neural networks~\cite{xiao2015learning, NIPS2017_7143,  patrini2016making}. Others have proposed more general classification frameworks that are not restricted to particular classifiers~\cite{natarajan2013learning, liu2016classification}. 

However, very little research has been conducted on solving the challenge posed by noisy labels in non-standard settings, where the magnitude of the noisy label problem is increased considerably. Some examples of such a non-standard setting occur for instance within image analysis~\cite{WU2018212}, document analysis~\cite{krithara2008semi}, named entity recognition~\cite{Ekbal2016}, crowdsourcing~\cite{nowak2010reliable}, or in the healthcare domain, used here as an illustrative case-study.
Non-standard settings include (i) \textit{Semi-supervised learning}~\cite{4787647} \cite{CHEN2017361}, referring to a situation where only a few (noisy) labeled data points are available, making the impact of noise in those few labels more prevalent,  and where information must also jointly be inferred from unlabeled data points. In healthcare, it may be realistic to obtain some labels through a (imperfect) manual labeling process, but the vast amount of data remains unlabeled; (ii) \textit{Multi-label learning}
~\cite{tahir2012multilabel, madjarov2012extensive, XU2013885, CHEN201661, LIU2018307,WANG20143405, trajdos2015extension, TRAJDOS201860, ZHUANG2018225}, wherein objects may not belong exclusively to one category. This situation occurs frequently in a number of domains, including healthcare, where for instance a patient could suffer from multiple chronic diseases; (iii) High-dimensional data, where the abundance of features and the limited (noisy) labeled data, lead to a curse of dimensionality problem. In such situations, \textit{dimensionality reduction} (DR)~\cite{Theodoridis:2008:PRF:1457541} is useful, either as a pre-processing step, or as an integral part of the learning procedure. This is a well-known challenge in health, where the number of patients in the populations under study frequently is small, but heterogeneous potential sources of data features from electronic health records for each patient may be enormous~\cite{lee2017medical, jensen2012mining, 7154395, Miotto}.  
 
In this paper, and to the best of our knowledge, we propose the first noisy label, semi-supervised and multi-label DR machine learning method, which we call the \emph{Noisy multi-label semi-supervised dimensionality reduction (NMLSDR)} method.
Towards that end, we propose a label propagation method that can deal with noisy multi-label data. Label propagation~\cite{zhu2002learning, zhu2003semi, yang2016revisiting, belkin2003using, 6409473, zhou2004learning, fan2017semi}, \cite{wang2016dynamic}, wherein one propagates the labels to the unlabeled data in order to obtain a fully labeled dataset, is one of the most successful and fundamental frameworks within semi-supervised learning. However, in contrast to many of these methods that clamp the labeled data, in our multi-label propagation method we allow the labeled part of the data to change labels during the propagation to account for noisy labels. 
In the second part of our algorithm we aim at learning a lower dimensional representation of the data by maximizing the feature-label dependence. Towards that end, similarly to other DR methods~\cite{zhang2010multilabel, XU2016172}, we employ the Hilbert-Schmidt independence criterion (HSIC)~\cite{gretton2005measuring}, which is a non-parametric measure of dependence.

The NMLSDR method is a DR method, which is general and can be used in many different settings, e.g. for visualization or as a pre-processing step before doing classification.
However, in order to test the quality of the NMLSDR embeddings, we (preferably) have to use some quantitative measures. 
For this purpose,  a common baseline classifier such as the multi-label k-nearest neighbor (ML-kNN) classifier~\cite{ZHANG20072038}  has been applied to the low-dimensional representations of the data~\cite{guo2016semi, 8059761}.
Even though this is a valid way to measure the quality of the embeddings, to apply a supervised classifier in a semi-supervised learning setting is not a realistic setup since one suddenly assumes that all labels are known (and correct).  Therefore, as an additional contribution,  we introduce a novel framework for semi-supervised classification of noisy multi-label data.

In our experiments, we compare NMLSDR to baseline methods on synthetic data, benchmark datasets, as well as a real-world case study, where we use it to identify the health status of patients suffering from potentially multiple  chronic diseases. 
The experiments demonstrate that for partially and noisy labeled multi-label data, NMLSDR is superior to existing DR methods according to seven different multi-label evaluation metrics and the Wilcoxon statistical test.

In summary, the contributions of the paper are as follows.
\begin{itemize}
    \item A new label noise-tolerant semi-supervised multi-label dimensionality reduction method based on dependence maximization.
    \item A novel framework for semi-supervised classification of noisy multi-label data.
    \item A comprehensive experimental section that illustrate the effectiveness of the NMLSDR on synthetic data, benchmark datasets and on a real-world case study.
\end{itemize}

The remainder of the paper is organized as follows. Related work is reviewed in Sec.~\ref{sec: related work}.
In Sec.~\ref{sec: NMLSDR}, we describe our proposed NMLSDR method and the novel  framework for semi-supervised classification of noisy multi-label data.
Sec.~\ref{sec:experiments} describes experiments on synthetic and benchmark datasets, whereas Sec.~\ref{sec: Case study} is devoted to the case study where we study chronically ill patients. We conclude the paper in Sec.~\ref{sec: Conclusion}.

\section{Related work}
\label{sec: related work}
In this section we review related unsupervised, semi-supervised and supervised DR methods.\footnote{DR may be obtained both by feature extraction, i.e. by a data transformation, and by feature selection~\cite{guyon2003introduction}. Here, we refer to DR in the sense of feature extraction.}

Unsupervised DR methods do not exploit label information and can therefore straightforwardly be applied to multi-label data by simply ignoring the labels. For example, principal component analysis (PCA)  aims to find the projection such that the variance of the input space is maximally preserved~\cite{jolliffe2011principal}. Other methods aim to find a lower dimensional embedding that preserves the manifold structure of the data, and examples of these include Locally linear embedding~\cite{roweis2000nonlinear}, Laplacian eigenmaps~\cite{belkin2002laplacian} and ISOMAP~\cite{tenenbaum2000global}.

One of the most well-known supervised DR methods is linear discriminative analysis (LDA)~\cite{fisher1936use}, which aims at finding the linear projection that maximizes the within-class similarity and at the same time minimizes the between-class similarity. 
LDA has been extended to multi-label LDA (MLDA) in several different ways~\cite{park2008applying, chen2007document, wang2010multi,lin2010mr, XU2018107}. The difference between these methods basically consists in the way the labels are weighted in the algorithm. Following the notation in~\cite{XU2018107},  wMLDAb~\cite{park2008applying} uses binary weights, wMLDAe~\cite{chen2007document} uses entropy-based weights, wMLDAc~\cite{wang2010multi} uses correlation-based weights, wMLDAf~\cite{lin2010mr} uses fuzzy-based weights, whereas wMLDAd~\cite{XU2018107} uses dependence-based weights.

Canonical correlation analysis (CCA)~\cite{hardoon2004canonical} is a method that maximizes the linear correlation between two sets of variables, which in the case of DR are the set of labels and the set of features derived from the projected space.  CCA can be directly applied also for multi-labels without any modifications.
Multi-label informed latent semantic indexing (MLSI)~\cite{yu2005multi} is a DR method that aims at both preserving the information of inputs and capturing the correlations between the labels.
In the Multi-label least square (ML-LS)  method one extracts a common subspace that is assumed to be shared among multiple labels by solving a generalized eigenvalue decomposition problem~\cite{ji2010shared}.

In~\cite{zhang2010multilabel}, a supervised method for doing DR based on dependence maximization~\cite{gretton2005measuring} called Multi-label dimensionality reduction via dependence maximization (MDDM) was introduced. MDDM attempts to maximize the feature-label dependence using the Hilbert-Schmidt independence criterion and was originally formulated in two different ways. MDDMp is based on orthonormal projection directions, whereas MDDMf makes the projected features orthonormal. Yu et al. showed that MDDMp can be formulated using least squares and added a PCA term to the cost function in a new method called Multi-label feature extraction via maximizing feature variance and feature-label dependence simultaneously (MVMD)~\cite{XU2016172}.

The most closely related existing DR methods to NMLSDR are the semi-supervised multi-label methods. The Semi-supervised dimension reduction for multi-label classification method (SSDR-MC)~\cite{qian2010semi}, Coupled dimensionality reduction and classification
for supervised and semi-supervised multilabel learning~\cite{GONEN2014132}, and Semisupervised 
multilabel learning with joint dimensionality reduction~\cite{yu2016semisupervised}  are semi-supervised multi-label methods that simultaneously learn a 
classifier and a low dimensional embedding.

Other semi-supervised multi-label DR methods are semi-supervised formulations of the corresponding supervised multi-label DR method. 
Blascho et al. introduced semi-supervised CCA based on Laplacian regularization~\cite{blaschko2011semi}.
Several different semi-supervised formulations of MLDA have also been proposed. 
Multi-label dimensionality reduction based on semi-supervised discriminant analysis (MSDA) adds two regularization terms computed from an adjacency matrix and a similarity correlation matrix, respectively,  to the MLDA objective function~\cite{li2010multi}.
In the Semi-supervised multi-label dimensionality reduction (SSMLDR)~\cite{guo2016semi} method one does label propagation to obtain soft labels for the unlabeled data. Thereafter the soft labels of all data are used to compute the MLDA scatter matrices. 
An other extension of MLDA is Semi-supervised multi-label linear discriminant analysis (SMLDA)~\cite{yu2017semi}, which later was modified and renamed
Semi-supervised multi-label dimensionality reduction based on dependence maximization (SMDRdm)~\cite{8059761}.  In SMDRdm the scatter matrices are computed based on only labeled data. However, a HSIC term is also added to the familiar Rayleigh quotient containing the two scatter matrices, which is computed based on soft labels for both labeled and unlabeled data obtained in a similar way as in SSMLDR.

Common to all these methods is that none of them explictly assume that the labels can be noisy. In SSMLDR and SMDRdm, the labeled data are clamped during the label propagation and hence cannot change. Moreover, these two methods are both based on LDA, which is known  heavily affected by outliers, and consequently also wrongly labeled data~\cite{HUBERT2004301, croux2001robust, hubert2008high}.

\section{The NMLSDR method}
\label{sec: NMLSDR}

We start this section by introducing notation and the setting for noisy multi-label semi-supervised linear feature extraction, and thereafter elaborate on our proposed NMLSDR method.

\subsection{Problem statement}
\label{sec: setup}

Let $\{ x_i \}_{i=1}^n$ be a set of $n$ $D$-dimensional data points, $x_i \in \mathbb{R}^D$. Assume that the data are ordered such that the $l$ first of the data points are labeled and $u$ are unlabeled, $l+u = n$. Let $X$ be a $n \times D$ matrix with the data points as row vectors.

Assume that the number of classes is $C$ and let $Y^L_{i}  \in \{0,1\}^C$ be the label-vector of data point $x_i$, $i = 1,\dots,l$. The elements are given by $Y^L_{ic} = 1 $, $c=1,\dots,C$  if data point $x_i$ belongs to the $c-$th class and $Y^L_{ic} = 0 $ otherwise.
Define the label matrix  $Y^L \in \{0,1\}^{l \times C}$ as the matrix with the known label-vectors $Y^L_{i}$, $i= 1,\dots,l$ as row vectors and let $Y^U \in \{0,1\}^{u \times C}$ be the corresponding label matrix of the unknown labels.

The objective of linear feature extraction is to learn a projection matrix $P \in \mathbb{R}^{D \times d} $ that maps a data point in the original feature space $x \in \mathbb{R}^D$ to a lower dimensional representation $z \in \mathbb{R}^d$,
\begin{equation}
    z = P^T x,
\end{equation}
where $d < D$ and $P^T$ denotes the transpose of the matrix $P$.

In our setting, we assume that the label matrix $Y^L$ is potentially noisy and that $Y^U$ is unknown. The first part of our proposed NMLSDR method consists of doing label propagation in order to learn the labels $Y^U$ and update the estimate of $Y^L$. We do this by introducing soft labels $F \in \mathbb{R}^{n \times C}$ for the label matrix $Y = \begin{pmatrix} Y^L \\ Y^U \end{pmatrix} $, where $F_{ic}$ represents the probability that data point $x_i$ belong to the $c-th$ class. We obtain $F$ with label propagation and thereafter use $F$ to learn the projection matrix $P$.
However, we start by explaining our label propagation method.

\subsection{Label propagation using a neighborhood graph} \label{sec: label propagation}

The underlying idea of label propagation is that similar data points should have similar labels. Typically, the labels are propagated using a neighborhood graph~\cite{zhu2002learning}. 
Here, inspired by~\cite{nie2010general}, we formulate a label propagation method for multi-labels that is robust to noise. The method is as follows.

\emph{Step 1.}
First, a neighbourhood graph is constructed. The graph is described by its adjacency matrix $W$, 
which can be designed e.g. by setting the entries to 
\begin{equation} \label{eq: W_ij exp}
    W_{ij} = \exp (-\sigma^{-2} \| x_i - x_j \|^2 ),
\end{equation}
where $\| x_i - x_j \|$ is the Euclidean distance between the datapoints $x_i$ and $x_j$, and $\sigma $ is a hyperparameter. Alternatively, one can use the Euclidian distance to compute a k-nearest neighbors (kNN) graph where the entries of $W$ are given by
\begin{equation}  \label{eq: W_ij binary}
    W_{ij} = 
    \begin{cases}
        1,  \quad \text{if $x_i$ among $x_j$'s $k$NN or $x_j$ among $x_i$'s $k$NN }\\
        0, \quad \text{otherwise}.\\
    \end{cases}
\end{equation}

\emph{Step 2.}
Symmetrically normalize the adjacency matrix $W$ by letting 
\begin{equation}
    \tilde{W} = D^{-1/2} W D^{-1/2},
\end{equation}
where $D$ is a diagonal matrix with entries given by $d_{ii} = \sum_{k=1}^n W_{ik}$.

\emph{Step 3.}
Calculate the stochastic matrix
\begin{equation}
    T = \tilde{D}^{-1} \tilde{W},
\end{equation}
where $\tilde{d}_{ii} = \sum_{k=1}^n \tilde{W}_{ik}$.
The entry $T_{ij}$ can now be considered as the probability of a transition from node $i$ to node $j$ along the edge between them. 

\emph{Step 4.}
Compute soft labels $F \in \mathbb{R}^{n \times C}$ by  iteratively using the following update rule
\begin{equation} \label{eq: F update}
    F(t+1) = I_{\alpha} T F(t) + (I-I_{\alpha})Y,
\end{equation}
where $I_{\alpha} $ is a $n \times n$ diagonal matrix with the hyperparameters $\alpha_i$, $0 \leq \alpha_i < 1$,  on the diagonal. To initialize $F$, we let $F(0) = Y$, where the unlabeled data are set to  $Y^U_{ic} = 0$, $c=1,\dots,C$.

\subsubsection{Discussion}

Setting $\alpha_i = 0$ for the labeled part of the data corresponds to clamping of the labels. However, this is not what we aim for in the presence of noisy labels. Therefore, a crucial property of the proposed framework is to set $\alpha_i > 0$ such that the labeled data can change labels during the propagation.

Moreover, we note that our extension of label propagation to multi-labels is very similar to the single-label variant introduced in~\cite{nie2010general}, with the exception that we do not add the outlier class, which is not needed in our case. In other extensions to the multi-label label propagation \cite{guo2016semi, 8059761},   the label matrix $Y$ is normalized such that the rows sum to 1, which ensures that the output of the algorithm $F$ also has rows that sum to 1. In the single-label case this makes sense in order to maintain the interpretability of probabilities. However, in the multi-label case the data points do not necessarily exclusively belong to a single class. Hence, the requirement $\sum_c F_{ic} = 1 $ does not make sense since then $x_i$ can maximally belong to one class if one think of $F$ as a probability and require the probability to be 0.5 or higher in order to belong to a class.

On the other hand, in our case, a simple calculation shows that  $ 0 \leq F_{ic}(t+1) \leq 1 $:
\begin{align}
    F_{ic}(t+1) &= \alpha_i \sum\limits_{m=1}^n T_{im} F_{mc}(t) + (1-\alpha_i) Y_{ic} \nonumber \\
    &\leq \alpha_i \sum\limits_{m=1}^n T_{im}  + (1-\alpha_i) = \alpha_i  + (1-\alpha_i) = 1,
\end{align}
since $F_{ic}(t) \leq 1$ and  $Y_{ic} \leq 1$. However, we do not necessarily have that $\sum_c F_{ic} = 1 $.

From matrix theory it is known that, given that $I - I_{\alpha}T$ is nonsingular, the solution of the linear iterative process~\eqref{eq: F update} converges to the solution of 
\begin{equation} \label{eq: F}
    (I - I_{\alpha}T) F = (I - I_{\alpha}) Y,
\end{equation}
for any initialization $F(0)$ if and only if $I_{\alpha}T$ is a \emph{convergent matrix}~\cite{meyer1977convergent} (spectral radius $\rho(I_{\alpha}T) < 1$).  $I_{\alpha}T$ is obviously convergent if $0 \leq \alpha_i < 1$ $\forall i$. Hence, we can find the soft labels $F$ by solving the linear system given by Eq.~\eqref{eq: F}.

Moreover, $F_{ic}$ can be interpreted as the probability that datapoint $x_i$ belongs to class $c$, and therefore, if one is interested in hard label assignments, $\tilde{Y}$, these can be found by letting $\tilde{Y}_{ic} = 1$ if $F_{ic} > 0.5$ and $\tilde{Y}_{ic} = 0$  otherwise.

\subsection{Dimensionality reduction via dependence maximization}
\label{sec:  dep max}

In this section we explain how we use the labels obtained using label propagation to learn the projection matrix $P$.

The motivation behind dependence maximization is that there should be a relation between the features and the label of an object. This should be the case also in the projected space. Hence, one should try to maximize the dependence between the feature similarity in the projected space and the label similarity. A common measure of such dependence is the Hilbert-Schmidt independence criterion (HSIC)~\cite{gretton2005measuring}, defined by
\begin{equation} \label{eq: HSIC}
    HSIC(X, Y) = \frac{1}{(n-1)^2} tr(K H L H),
\end{equation}
where $tr$ denotes the trace of a matrix. $H \in \mathbb{R}^{n \times n}$ is given by $H_{ij} = \delta_{ij} - n^{-1} $, where $\delta_{ij}=1$ if $i=j$, and $\delta_{ij}=0$ otherwise. 
$K$ is a kernel matrix over the feature space, whereas  $L$ is a kernel computed over the label space.

Let the projection of $x$ be given by the projection matrix $P \in \mathbb{R}^{D \times d} $ and function $\Phi: \: \mathbb{R}^D \to \mathbb{R}^d $, $\Phi(x) = P^T x $. We select a linear kernel over the feature space, and therefore the kernel function is given by
\begin{equation}
    \mathcal{K}(x_i, x_j) = \langle \Phi(x_i), \Phi(x_j) \rangle  = \langle P^T x_i, P^T x_j  \rangle =  P^T x_i  x_j^T P
\end{equation}
Hence, given data $\{ x_i \}_{i=1}^n $, the kernel matrix can be approximated by $K = X P^T P X^T$.

The kernel over the label space, $\mathcal{L}$, is given via the labels $y_i \in \{0, 1\}^C$. One possible such kernel is the linear kernel
\begin{equation}
    \mathcal{L}(y_i, y_j) = \langle y_i, y_j \rangle.
\end{equation}
However, in our semi-supervised setting, some of the labels are unknown and some are noisy. Hence, the kernel $ \mathcal{L} $ cannot be computed. In order to enable DR in our non-standard problem, we propose to estimate the kernel using the labels obtained via our label propagation method. 
For the part of the data that was labeled from the beginning we use the hard labels, $\tilde{Y}^L$, obtained from the label propagation, whereas for the unlabeled part we use the soft labels, $F^U$. 
Hence, the kernel is approximated via 
$L = \tilde{F} \tilde{F}^T$,  
where $\tilde{F}  = \begin{pmatrix} \tilde{Y}^L \\ F^U \end{pmatrix} $.
The reason for using the hard labels obtained from label propagation for the labeled part is that we want some degree of certainty for those labels that change during the propagation (if the soft label $F^L_{ic}$ changes with less than 0.5 from its initial value 0 or 1 during the propagation, the hard label $Y^L_{ic}$ does not change).

The constant term, $(n-1)^{-2}$, in Eq.~\eqref{eq: HSIC} is irrelevant in an optimization setting. Hence, by inserting the estimates of the kernels into Eq.~\eqref{eq: HSIC}, the following objective function is obtained,
\begin{equation}
    \Psi(P) =    tr(H X P^T P X^T H \tilde{F}  \tilde{F}^T) =  tr(P^T X^T H  \tilde{F}  \tilde{F}^T H X P).
\end{equation}
Note that the matrix $ X^T H \tilde{F}  \tilde{F}^T H X$ is symmetric. Hence, by requiring that the projection directions are orthogonal and that the new dimensionality is $d$,  the following optimization problem is obtained
\begin{align} \label{eq: arg max}
   \argmax_{P} \Psi(P) &= \argmax_{P} tr(P^T (X^T H \tilde{F}  \tilde{F}^T H X) P), \\
    &s.t. \: P \in \mathbb{R}^{D \times d}, \: P P^T = I. \nonumber
\end{align}
As a consequence of the Courant-Fisher characterization~\cite{saad1992numerical}, it follows that the maximum is achieved when $P$ is an orthonormal basis corresponding to the $ d$ largest eigenvalues. Hence, $P$ can be found by solving the eigenvalue problem
\begin{equation}
    X^T H \tilde{F}  \tilde{F}^T H X P = \Lambda P.
\end{equation}

The dimensionality of the projected space, $d$, is upper bounded by the rank of $\tilde{F}  \tilde{F}^T$, which in turn is upper bounded by the number of classes $C$. Hence, $d$ cannot be set larger than $C$. 
The pseudo-code of the NMLSDR method is shown in Alg.~\ref{alg:algorithm}.

\begin{algorithm}
\small
\caption{Pseudo-code for NMLSDR}
\label{alg:algorithm}
\begin{algorithmic}[1]
\Require $X \: : \: n \times D$ feature matrix, $Y \: : \: n \times C$ label matrix,  hyperparameters $k$, $I_{\alpha}$ and $d$.
\State Initialize $F$ by letting $F(0) = Y$, where the unlabeled data are set to  $Y^U_{ic} = 0$, $c=1,\dots,C$. 
\State 
Construct a neighbourhood graph by calculating the adjacency matrix $W$ using Eq.~(2) or~(3).
\State 
Symmetrically normalize the adjacency matrix $W$ by letting 
$
    \tilde{W} = D^{-1/2} W D^{-1/2},
$
\State
Calculate the stochastic matrix
$
    T = \tilde{D}^{-1} \tilde{W},
$
where $\tilde{d}_{ii} = \sum_{k=1}^n \tilde{W}_{ik}$.
\State 
 Solve the linear system $ (I - I_{\alpha}T) F = (I - I_{\alpha}) Y$.
\State Compute $ \tilde{F} $.
\State Construct the  matrix $ X^T H \tilde{F}  \tilde{F}^T H X$.
\State Eigendecompose $ X^T H \tilde{F}  \tilde{F}^T H X$ and construct projection matrix $P \in \mathbb{R}^{D \times d}$.
\Ensure Projection $P \: : \: \mathbb{R}^D \to \mathbb{R}^d$ 
\end{algorithmic}
\end{algorithm}

\subsection{Semi-supervised classification for noisy multi-label data}
\label{sec: End-to-end framework}

The multi-label k-nearest neighbor (ML-kNN) classifier~\cite{ZHANG20072038} is a widely adopted classifier for multi-label classification. However, similarly to many other classifiers, its performance can be hampered if the dimensionality of the data is too high. Moreover, the ML-kNN classifier only works in a completely supervised setting. To resolve these problems, as an additional contribution of this work,  we introduce a novel  framework for semi-supervised classification of noisy multi-label data, consisting of two steps. 
In the first step, we compute a low dimensional embedding using NMLSDR. The second step consists of applying a semi-supervised ML-kNN classifier.
For this classifier we use our label propagation method on the learned embedding to obtain a fully labeled dataset, and thereafter apply the ML-kNN classifier.

\section{Experiments}
\label{sec:experiments}

In this paper, we have proposed a method for computing a low-dimensional embedding of noisy, partially labeled multi-label data. However, it is not a  straightforward task to measure how well the method works. Even though the method is definitely relevant to real-world problems (illustrated in the case study in Sec.~\ref{sec: Case study}), the framework cannot be directly applied to most multi-label benchmark datasets since most of them are completely labeled, and the labels are assumed to be clean. 
Moreover,  the NMLSDR provides a low dimensional embedding of the data, and we need a way to measure how good the embedding is. If the dimensionality is 2 or 3, this can to some degree be done visually by plotting the embedding.
However, in order to quantitatively measure the quality and simultaneously  maintain a realistic setup, we will apply our proposed end-to-end framework for semi-supervised classification and dimensionality reduction. In our experiments, this realistic semi-supervised setup will be applied in an illustrative example on synthetic data and in the case study.

A potential disadvantage of using a semi-supervised classifier, is that it does not necessarily isolate effect of the DR method that is used to compute the embedding. For this reason, we will also test our method on some benchmark datasets, but in order to keep everything coherent, except for the method used to compute the embedding, we  compute the embedding using NMLSDR and baseline DR methods based on only the noisy and partially labeled multi-label training data. Thereafter, we assume that the true multi-labels are available when we train the ML-kNN classifier on the embeddings.

The remainder of this section is organized as follows. First we describe the performance measures we employed, baseline DR methods, and how we select hyper-parameters. Thereafter we provide an illustrative example on synthetic data, and secondly experiments on the benchmark data. The case study is described in the next section.

\subsection{Evaluation metrics} \label{sec: evaluation metrics}
Evaluation of performance is more complicated in a multi-label setting than for traditional single-labels. In this work, we decide use the seven different evaluation criteria that were employed in~\cite{zhang2010multilabel}, namely Hamming loss (HL), Macro F1-score (MaF1),  Micro F1 (MiF1), Ranking loss (RL), Average precision (AP), One-error (OE)  and  Coverage (Cov).

HL simply evaluates the number of times there is a mismatch between the predicted label and the true label, i.e.
\begin{equation}
    HL = \sum\limits_{i=1}^n \frac{ \| \hat{y}_i \oplus y_i \|_1 }{n C},
\end{equation}
where $\hat{y}_i$  denotes the predicted label vector of data point $x_i$ and $ \oplus$ is the XOR-operator. MaF1 is obtained by first computing the F1-score for each label, and then averaging over all labels.
\begin{equation}
    MaF1 = \frac{1}{C} \sum\limits_{c=1}^C \frac{ 2  \sum_{i=1}^n \hat{y}_{ic} y_{ic} }{\sum_{i=1}^n \hat{y}_{ic} + \sum_{i=1}^n y_{ic} },
\end{equation}
MiF1 calculates the F1 score on the predictions of different labels as a whole,
\begin{equation}
    MiF1 = \frac{ 2  \sum_{i=1}^n \sum_{c=1}^C \hat{y}_{ic} y_{ic} }{\sum_{i=1}^n \sum_{c=1}^C \hat{y}_{ic} + \sum_{i=1}^n \sum_{c=1}^C y_{ic} },
\end{equation}
We note that HL, MiF1 and MaF1 are computed based on hard labels assignments, whereas the four other measures are computed based on soft labels. In all of our experiments, we obtain the hard labels by putting a threshold at 0.5.  

RL computes the average ratio of reversely ordered label pairs of each data point. AP evaluates the average fraction of relevant labels ranked higher than a particular relevant label. OE gives the ratio of data points where the most confident predicted label is wrong. Cov gives an average of how far one needs to go down on the list of ranked labels to cover all the relevant labels of the data point. For a more detailed description of these measures, we point the interested reader to~\cite{wu2016unified}.

In this work, we modify four of the evaluation metrics such that all of them take values in the interval $[0, 1]$ and ``higher always is better". Hence, we define
\begin{align}
    HL' &= 1 - HL, \\
    RL'  &= 1 - RL, \\
    OE' &=1 -   OE, 
\end{align}
and normalized coverage (Cov') by
\begin{equation}
    Cov' = 1- Cov/(C-1).
\end{equation}

\subsection{Baseline dimensionality reduction methods} In this work, we consider the following other DR methods: CCA, MVMD, MDDMp, MDDMf and four variants of MLDA, namely wMLDAb, wMLDAe, wMLDAc and wMLDAd. These methods are supervised and require labeled data, and are therefore trained only on the labeled part of the training data. In addition, we compare to a semi-supervised method, SSMLDR, which we adapt to noisy multi-labels by using  the label propagation algorithm we propose in this paper instead of the label propagation method that was originally proposed in SSMLDR. We note that the computational complexity of NMLSDR and the all the baselines is of the same order as all of them require a step involving eigendecomposition.

\subsection{Hyper-parameter selection and implementation settings}
For the ML-kNN classifier we set $k=10$. The effect of varying the number of neighbors will be left for further work.
In order to learn the NMLSDR embedding we use a kNN-graph with $k =10$ and binary weights. Moreover, we set $\alpha_i = 0.6 $ for labeled data and $\alpha_i = 0.999$ for unlabeled data. By doing so, one ensures that an unlabeled datapoint is not affected by its initial value, but gets all contribution from the neighbors during the propagation. 
All experiments are run in Matlab using an Ubuntu 16.04 64-bit system with 16 GB RAM and an Intel Core  i7-7500U processor.

\subsection{Illustrative example on synthetic toy data} \label{sec: illustrative example}

\paragraph{Dataset description}
To test the framework in a controlled experiment, a synthetic dataset is created as follows.

A dataset of size 8000 samples is created, where each of the data points has dimensionality  320. The number of classes is set to 4, and we generate 2000 samples from each class. 30\% from class 1 also belong to class 2, and vice versa.  20\% from class 2 also belong to class 3 and vice versa, whereas 25\% from class 3 also belong to class 4 and vice versa.

A sample from class $i$ is generated by randomly letting 10\% of the features in the interval $\{ 20(i-1)+1, \dots, 20 i\} $ take a random integer value between 1 and 10. Since there are 4 classes, this means that the first 80 features are directly dependent on the class-membership. 

For the remaining 240 features we consider 20 of them at the time. We randomly select 50\% of the 8000 samples and randomly let 20\% of the 20 features take a random integer value between 1 and 10.
We repeat this procedure for the 12 different sets of 20 features $\{ 20(i-1)+1, \dots, 20 i\} $, $i = 5,6,\dots, 16$.

All features that are not given a value using the procedure described above are set to 0. Noise is injected into the labels by randomly flipping a fraction $p = 0.1$  of the labels and we make the data partially labeled by removing 50 \% of the labels. 
2000 of the samples are kept aside as an independent test set.
We note that noisy labels are often easier and cheaper to obtain than true labels and it is therefore not unreasonable that the fraction of labeled examples is larger than what it commonly is in traditional semi-supervised learning settings.

\paragraph{Results}

We apply the NMLSDR method in combination with the semi-supervised ML-kNN classifier as explained above and compare to SSMLDR. We create two baselines by, for both of these methods, using a  different value for the hyperparameter $\alpha_i$ for the labeled part of the data, namely 0, which corresponds to clamping. We denote these two baselines by SSMLDR* and NMLSDR*. In addition, we compare to baselines that only utilize the labeled part of the data, namely the supervised DR methods explained above in combination with a ML-kNN classifier. The data is standardized to 0 mean and 1 in standard deviation and we let the dimensionality of the embedding be 3.

\begin{figure}[t]
    \centering
        \begin{subfigure}{.49\textwidth}
            \includegraphics[trim = 42mm 89mm 44mm 87mm, clip,height = 5.0cm]{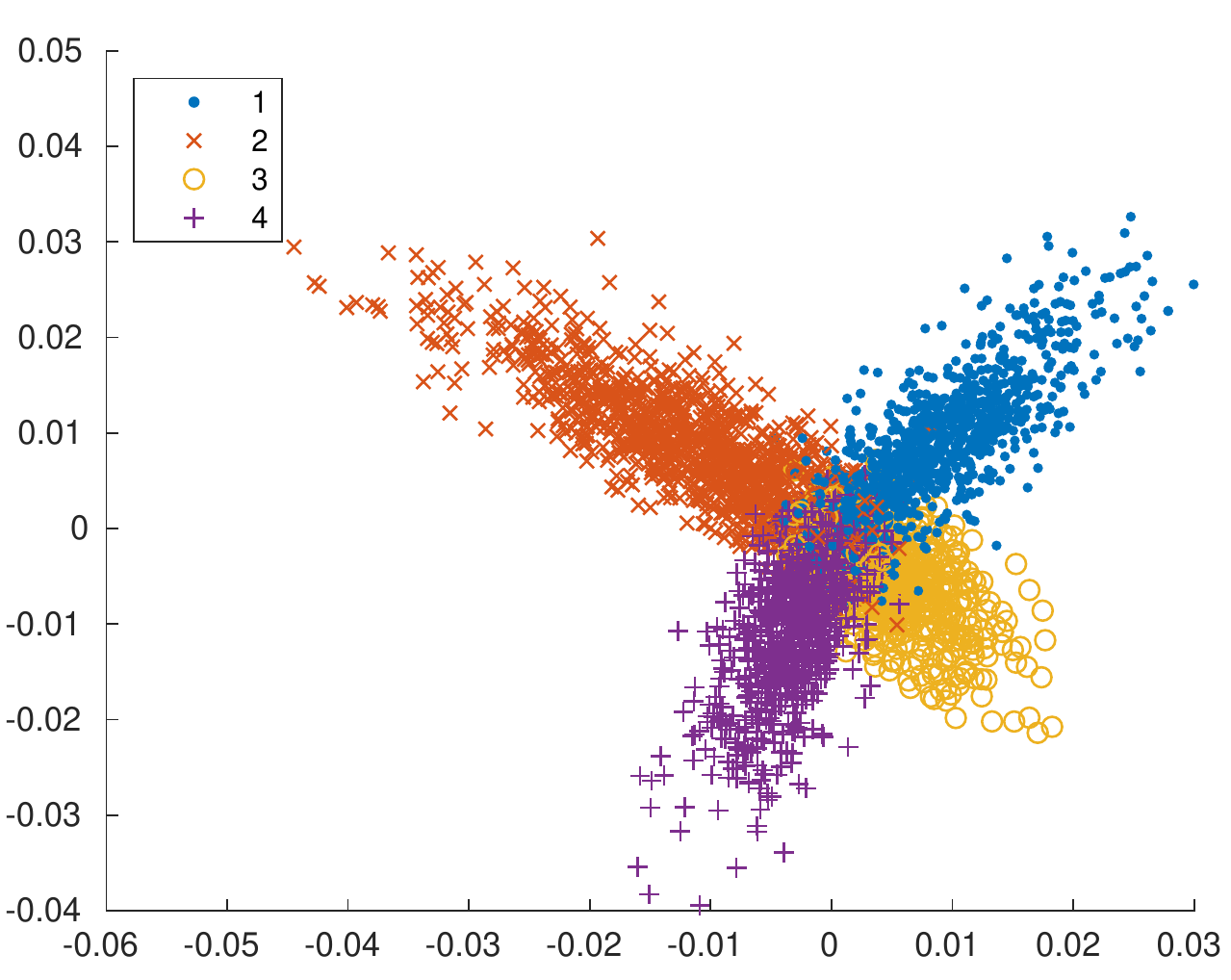}
            \caption{}
            \label{fig:SMLDR_synt}
        \end{subfigure} 
         \begin{subfigure}{.49\textwidth}
            \includegraphics[trim = 42mm 89mm 44mm 87mm, clip,height = 5.0cm]{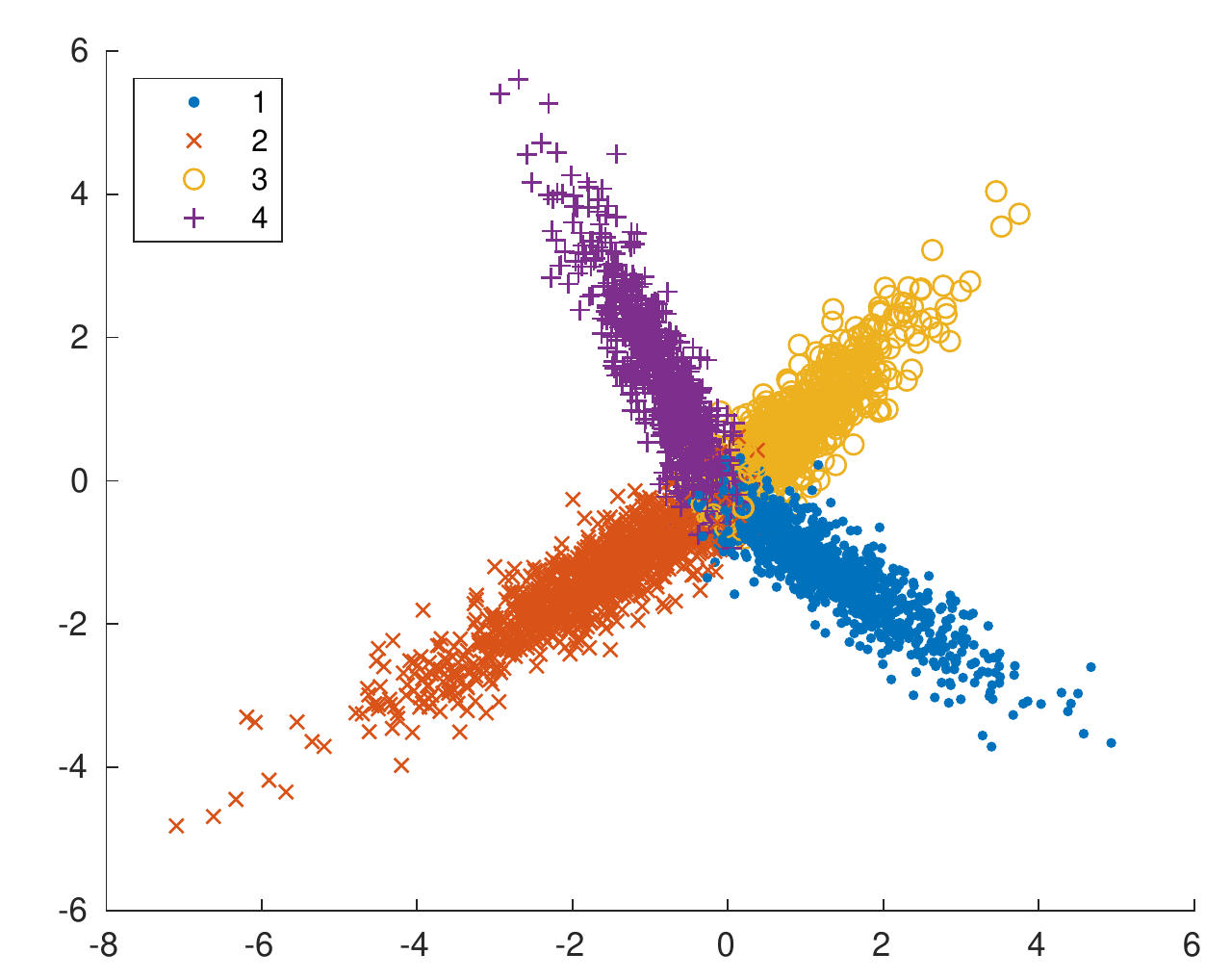}
            \caption{}
            \label{fig:SMDDM_synt}
        \end{subfigure}
        \begin{subfigure}{.49\textwidth}
            \includegraphics[trim = 42mm 89mm 44mm 87mm, clip,height = 5.0cm]{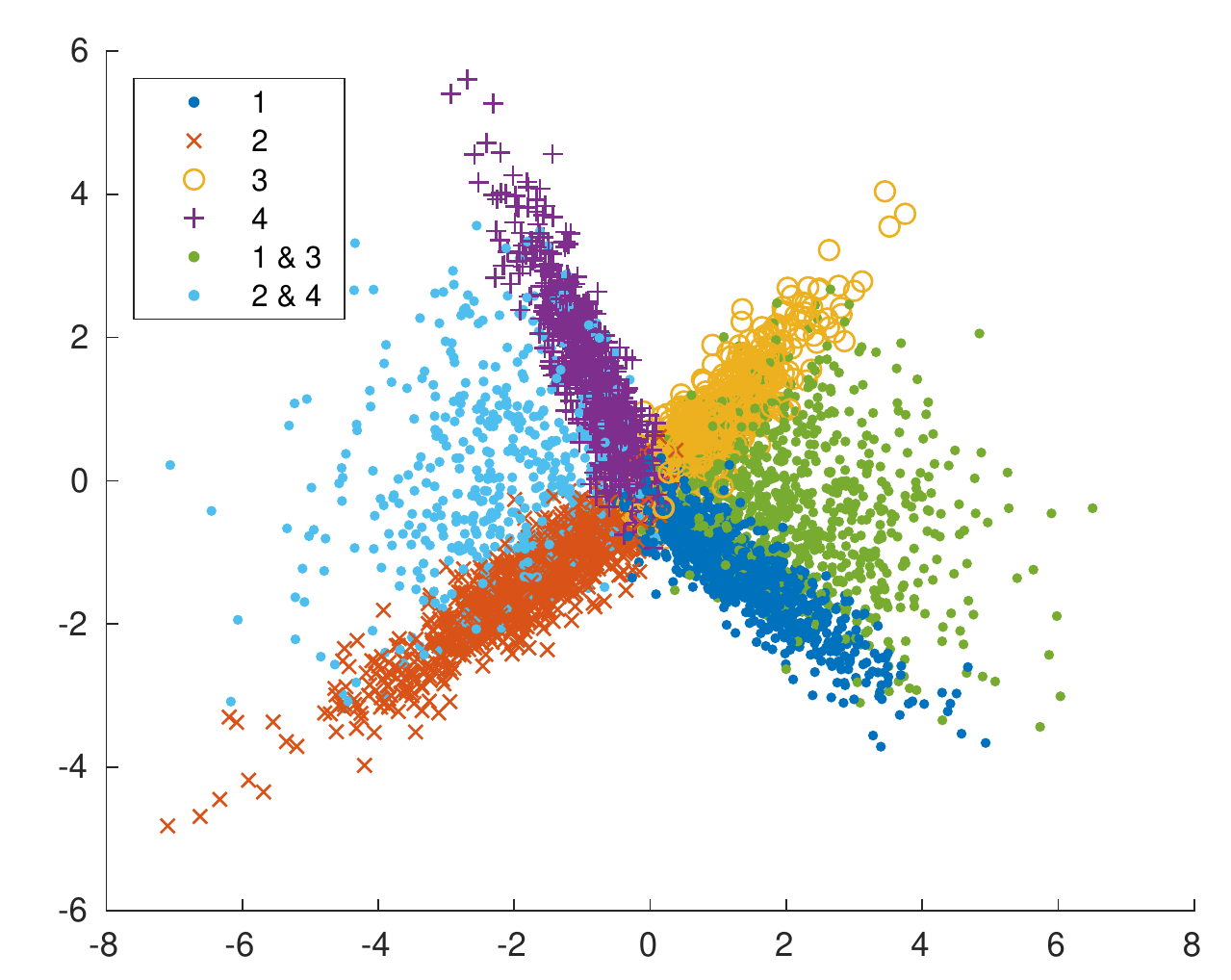}
            \caption{}
            \label{fig:SMDDM_full_synt}
        \end{subfigure}
         \begin{subfigure}{.49\textwidth}
            \includegraphics[trim = 42mm 89mm 44mm 87mm, clip,height = 5.0cm]{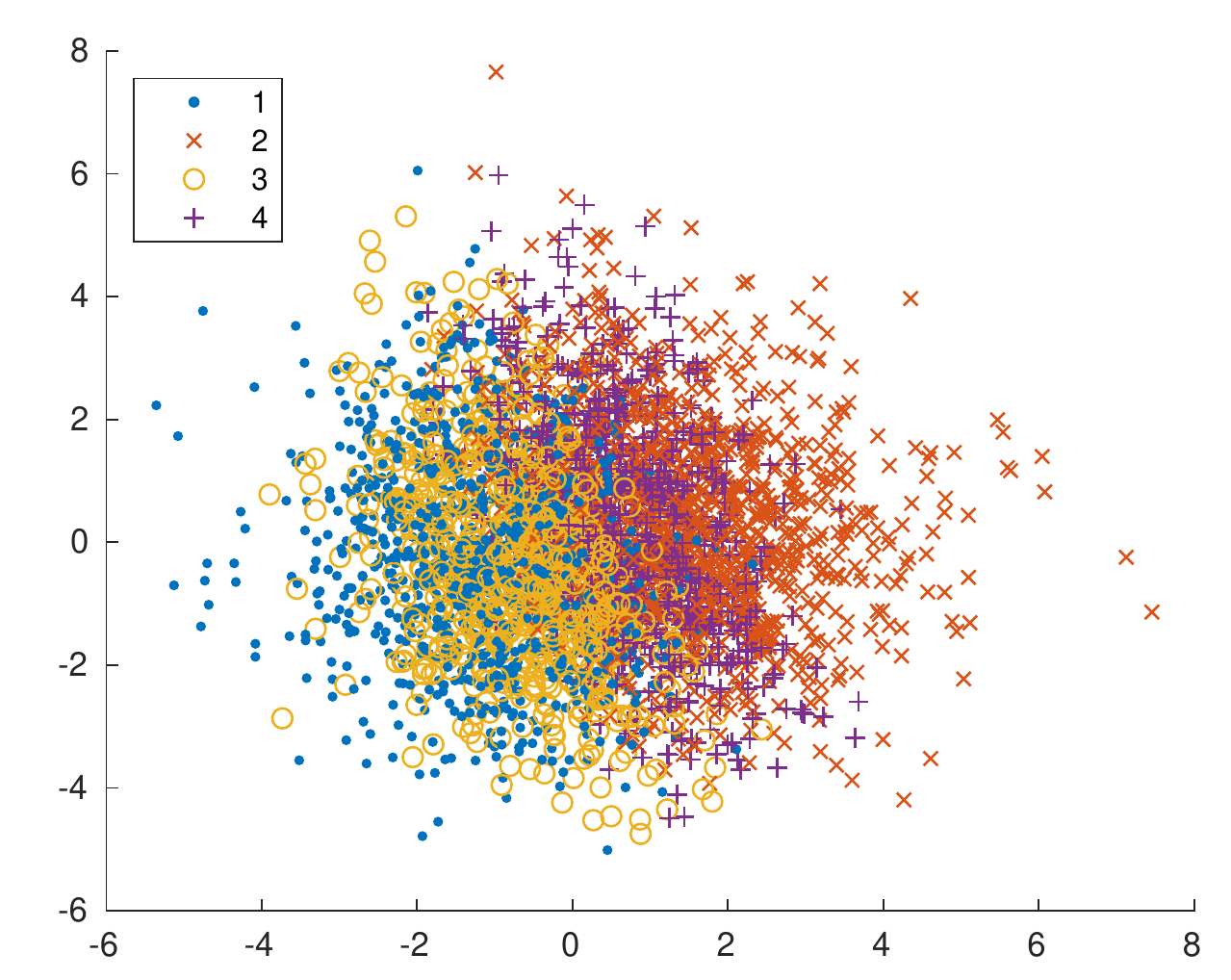} 
            \caption{}
            \label{fig:pca_synt}
        \end{subfigure}  
    \caption{3 dimensional embedding of the synthetic dataset obtained using (a) SSMLDR; (b)  NMLSDR; (c) NMLSDR with multi-classes included; and (d) PCA.}
\label{fig:synt}
\end{figure}

Fig.~\ref{fig:SMLDR_synt} and \ref{fig:SMDDM_synt} show the embeddings obtained obtained using SSMLDR and NMLSDR, respectively. For ivisualization purposes, we have only plotted those datapoints that exclusively belong to one class. In Fig.~\ref{fig:SMDDM_full_synt}, we have added two of the multi-classes for the NMLSDR embedding. For comparison, we also added the embedding obtained using PCA in Fig.~\ref{fig:pca_synt}. As we can see, in the PCA embedding the classes are not separated from each other, whereas in the NMLSDR and SSMLDR embeddings the classes are aligned along different axes. It can be seen that the classes are better separated and more compact in the NMLSDR embedding than the SSMLDR embedding. Fig.~\ref{fig:SMDDM_full_synt} shows that the data points that belong to multiple classes are placed where they naturally belong, namely between the axes corresponding to both of the classes they are member of.

\begin{table}[t]
    \centering
    \scriptsize{
    \begin{tabular}{@{ }lccccccc@{ }}
    \hline
 Method      & HL'  & RL' & AP & OE' & Cov' & MaF1 & MiF1 \\
     \hline
CCA&0.863 & 0.884 & 0.898 & 0.852 & 0.816 & 0.787 & 0.785  \\ 
MVMD&0.906 & 0.912 & 0.924 & 0.897 & 0.836 & 0.850 & 0.849  \\ 
MDDMp&0.906 & 0.911 & 0.924 & 0.897 & 0.836 & 0.851 & 0.850  \\ 
MDDMf&0.859 & 0.888 & 0.900 & 0.855 & 0.819 & 0.785 & 0.783 \\ 
wMLDAb&0.844 & 0.871 & 0.885 & 0.831 & 0.807 & 0.754 & 0.750  \\ 
wMLDAe&0.864 & 0.885 & 0.899 & 0.855 & 0.818 & 0.790 & 0.788  \\ 
wMLDAc&0.865 & 0.887 & 0.900 & 0.857 & 0.818 & 0.787 & 0.785  \\ 
wMLDAd&0.869 & 0.891 & 0.907 & 0.869 & 0.822 & 0.788 & 0.786 \\  
SSMLDR* & 0.863 & 0.883 & 0.899 & 0.859 & 0.814 & 0.796 & 0.793  \\
SSMLDR &0.879 & 0.898 & 0.910 & 0.871 & 0.827 & 0.817 & 0.814  \\ 
NMLSDR* & 0.907 & 0.919 & 0.929 & 0.903 & 0.842 & 0.861 & 0.859 \\ 
NMLSDR & \textbf{0.913} & \textbf{0.925} & \textbf{0.935} & \textbf{0.912} & \textbf{0.846} & \textbf{0.868} & \textbf{0.866}  \\ 
     \hline  
    \end{tabular}
    }
    \caption{Results in terms of 7 metrics for the synthetic dataset.}
    \label{tab:classication_synt}
\end{table}

\begin{table}[t]
    \centering
     \scriptsize{
    \begin{tabular}{lllllll}
    \hline
       Dataset  &  Domain &  Train instances &  Test instances &  Attributes &  Labels & Cardinality \\
        \hline 
        Birds & audio & 322 & 323 & 260 & 19 & 1.06 \\
        Corel &  scene & 5188 & 1744 & 500 & 153 & 2.87 \\
        Emotions & music & 391 & 202 & 72 & 6 & 1.81 \\
        Enron & text & 1123 & 579 & 1001 & 52 & 3.38 \\
        Genbase & biology & 463 & 199 & 99 & 25 & 1.26 \\
        Medical & text & 645 & 333 & 1161 & 39 & 1.24  \\
        Scene & scene & 1211 & 1196 & 294 & 6 & 1.06  \\
        Tmc2007 & text & 3000 & 7077 & 493 & 22 & 2.25 \\
        Toy & synthetic & 6000 & 2000 & 320 & 4 & 1.38 \\
        Yeast & biology & 1500 & 917 & 103 & 14 & 4.23 \\
        \hline
    \end{tabular}}
    \caption{Description of benchmark datasets considered in our experiments.}
    \label{tab:datasets}
\end{table}

Tab.~\ref{tab:classication_synt} shows the results obtained using the different methods on the synthetic dataset. As we can see, our proposed method gives the best performance for all metrics.  Moreover, NMLSDR with $\alpha_i^L = 0$, which corresponds to clamping of the labeled data during label propagation gives the second best results but cannot compete with our proposed method, in which the labels are allowed to change during the propagation to account for noisy labels.
We also note that, even though the SSMLDR improves the MLDA approaches that are based on only the labeled part of the data, it gives results that are considerably worse than NMLSDR.

\subsection{Benchmark datasets}

\paragraph{Experimental setup}
We consider the following benchmark datasets~\footnote{Downloaded from mulan.sourceforge.net/datasets-mlc.html}: Birds, Corel, Emotions, Enron, Genbase, Medical, Scene, Tmc2007 and Yeast. We also add our synthetic toy dataset as a one of our benchmark datasets (described in Sec.~\ref{sec: illustrative example}). These datasets are shown in Tab.~\ref{tab:datasets}, along with some useful characteristics.  In order to be able to apply our framework to the benchmark datasets, we randomly flip 10 \% of the labels to generate noisy labels and let 30 \% of the data points training sets be labeled. All datasets are standardized to zero mean and standard deviation one.

We apply the DR methods to the partially and noisy labeled multi-label training sets in order to learn the projection matrix $P$, which in turn is used to map the D-dimensional training and test sets to a $d-$dimensional representation. $d$ is set as large as possible, i.e. to $C-1 $ for the MLDA-based methods and $C$ for the other methods. Then we train a ML-kNN classifier using the low-dimensional training sets, assuming that the true multi-labels are known and validate the performance on the low-dimensional test sets.

In total we are evaluating the performance over 10 different datasets and across 7 different performance measures for all the feature extraction methods we use. Hence, to investigate which method performs better according to the different metrics, we also report the number of times each method gets the highest value of each metric. In addition, we compare all pairs of methods by using a Wilcoxon signed rank test with 5\% significance level~\cite{demvsar2006statistical}. Similarly to~\cite{XU2018107}, if method A performs better than B according to the test, A is assigned the score 1 and B the score 0. If the null hypothesis (method A and B perform equally) is not rejected, both A and B are assigned an equal score of 0.5.

\paragraph{Results}

\begin{table}[!t]
    \centering
    \scriptsize{
    \begin{tabular}{@{\:}l@{\:}cc@{\:\:\:}c@{\:\:\:}c@{\:\:\:}c@{\:\:\:}c@{\:\:\:}c@{\:\:\:}c@{\:\:\:}c@{\:\:\:}c@{\:}}
    \hline
        & CCA  & MVMD & MDDMp & MDDMf & wMLDAb & wMLDAe & wMLDAc  & wMLDAd & SSMLDR & NMLSDR \\
        \hline
Birds& 0.947& 0.950& 0.950& 0.947& 0.948& 0.949& 0.949& 0.949& 0.949& \textbf{0.951} \\ 
Corel& \textbf{0.980}& \textbf{0.980}& \textbf{0.980}& \textbf{0.980}& \textbf{0.980}& \textbf{0.980}& \textbf{0.980}& \textbf{0.980}& \textbf{0.980}& \textbf{0.980} \\ 
Emotions& 0.715& 0.771& 0.778& 0.711& 0.696& 0.714& 0.709& 0.717& 0.786& \textbf{0.787} \\ 
Enron& 0.941& \textbf{0.950}& \textbf{0.950}& 0.942& 0.941& 0.941& 0.941& 0.940& 0.938& \textbf{0.950} \\ 
Genbase& 0.989& 0.996& 0.996& 0.988& 0.990& 0.991& 0.988& 0.989& 0.994&\textbf{0.997} \\ 
Medical& \textbf{0.976}& 0.974& 0.974& \textbf{0.976}& 0.974& 0.975& 0.975& \textbf{0.976}& 0.966& 0.975 \\ 
Scene& 0.810& 0.899& \textbf{0.900}& 0.809& 0.810& 0.814& 0.817& 0.810& 0.873& 0.897 \\ 
Tmc2007& 0.914& 0.928& 0.928& 0.912& 0.911& 0.911& 0.911& 0.916& 0.922& \textbf{0.929} \\ 
Toy& 0.836& 0.894& 0.894& 0.839& 0.821& 0.831& 0.831& 0.854& 0.861& \textbf{0.903} \\ 
Yeast& 0.780& 0.791& 0.790& 0.782& 0.785& 0.783& 0.781& 0.781& \textbf{0.793}& \textbf{0.793} \\ 
\hline
\# Best values& 2& 2& 3& 2& 1& 1& 1& 2& 2& \textbf{8} \\ 
Wilcoxon& 2.0& 7.0& 7.5& 2.5& 2.0& 3.0& 2.5& 3.5& 6.0& \textbf{9.0} \\ 
     \hline  
    \end{tabular}
    }
    \caption{Performance in terms of 1 - Hamming loss (HL') across 10 different benchmark datasets.}
    \label{tab:HL}
\end{table}

\begin{table*}[!h]
    \centering
     \scriptsize{
    \begin{tabular}{@{\:}l@{\:}cc@{\:\:\:}c@{\:\:\:}c@{\:\:\:}c@{\:\:\:}c@{\:\:\:}c@{\:\:\:}c@{\:\:\:}c@{\:\:\:}c@{\:}}
    \hline
        & CCA  & MVMD & MDDMp & MDDMf & wMLDAb & wMLDAe & wMLDAc  & wMLDAd & SSMLDR & NMLSDR \\
        \hline
Birds& 0.715& 0.766& 0.767& 0.734& 0.709& 0.718& 0.719& 0.725& 0.681& \textbf{0.771} \\ 
Corel& 0.800& 0.808& 0.808& 0.800& 0.799& 0.799& 0.800& 0.800& 0.801& \textbf{0.814 }\\ 
Emotions& 0.695& 0.824& 0.824& 0.709& 0.693& 0.700& 0.676& 0.714& 0.829& \textbf{0.845} \\ 
Enron& 0.894& 0.911& 0.911& 0.893& 0.893& 0.892& 0.891& 0.893& 0.883& \textbf{0.914} \\ 
Genbase& 0.993& 0.995& 0.995& 0.993& 0.994& 0.992& 0.992& 0.991& 0.995& \textbf{1.000} \\ 
Medical& 0.925& \textbf{0.952}& 0.949& 0.925& 0.916& 0.921& 0.919& 0.945& 0.856& 0.946 \\ 
Scene& 0.585& \textbf{0.900}& 0.898& 0.629& 0.574& 0.583& 0.572& 0.616& 0.853& 0.898 \\ 
Tmc2007& 0.831& 0.906& 0.906& 0.830& 0.830& 0.830& 0.831& 0.847& 0.872& \textbf{0.910} \\ 
Toy& 0.871& 0.909& 0.909& 0.870& 0.849& 0.865& 0.861& 0.888& 0.887& \textbf{0.926 }\\ 
Yeast& 0.806& \textbf{0.820}& 0.819& 0.811& 0.810& 0.809& 0.806& 0.803& 0.818& 0.816 \\ 
\hline 
\# Best values& 0& 3& 0& 0& 0& 0& 0& 0& 0& \textbf{7} \\ 
Wilcoxon& 3.0& 8.0& 7.5& 4.5& 1.5& 2.0& 2.0& 5.0& 3.0& \textbf{8.5 }\\ 
     \hline  
    \end{tabular}
    }
    \caption{Performance in terms of 1 - Ranking loss (RL') across 10 different benchmark datasets.}
    \label{tab:RL}
\end{table*}

\begin{table*}[!h]
    \centering
    \scriptsize{
    \begin{tabular}{@{\:}l@{\:}cc@{\:\:\:}c@{\:\:\:}c@{\:\:\:}c@{\:\:\:}c@{\:\:\:}c@{\:\:\:}c@{\:\:\:}c@{\:\:\:}c@{\:}}
    \hline
        & CCA  & MVMD & MDDMp & MDDMf & wMLDAb & wMLDAe & wMLDAc  & wMLDAd & SSMLDR & NMLSDR \\
        \hline
Birds& 0.389& 0.499& 0.500& 0.426& 0.374& 0.392& 0.379& 0.424& 0.357& \textbf{0.502} \\ 
Corel& 0.260& 0.277& 0.277& 0.261& 0.265& 0.263& 0.263& 0.268& 0.266& \textbf{0.288} \\ 
Emotions& 0.669& 0.781& 0.773& 0.686& 0.672& 0.687& 0.666& 0.704& 0.799& \textbf{0.808} \\ 
Enron& 0.592& 0.669& 0.670& 0.583& 0.584& 0.582& 0.580& 0.578& 0.526& \textbf{0.675} \\ 
Genbase& 0.963& 0.990& 0.993& 0.964& 0.960& 0.968& 0.963& 0.969& 0.984& \textbf{0.997} \\ 
Medical& 0.673& 0.722& 0.716& 0.666& 0.644& 0.674& 0.669& 0.723& 0.446& \textbf{0.725} \\ 
Scene& 0.491& \textbf{0.836}& 0.835& 0.534& 0.481& 0.488& 0.475& 0.521& 0.781& 0.834 \\ 
Tmc2007& 0.584& 0.714& 0.713& 0.587& 0.579& 0.576& 0.577& 0.623& 0.662& \textbf{0.721} \\ 
Toy& 0.882& 0.921& 0.921& 0.880& 0.862& 0.880& 0.875& 0.900& 0.897& \textbf{0.933 }\\ 
Yeast& 0.732& \textbf{0.748}& 0.747& 0.731& 0.733& 0.733& 0.729& 0.725& 0.745& 0.741 \\ 
\hline
\# Best values& 0& 2& 0& 0& 0& 0& 0& 0& 0& \textbf{8} \\ 
Wilcoxon& 3.5& 7.5& 7.5& 4.0& 1.0& 3.5& 1.0& 5.0& 3.0& \textbf{9.0} \\ 
     \hline  
    \end{tabular}
    }
    \caption{Performance in terms of Average precision (AP) across 10 different benchmark datasets.}
    \label{tab:AP}
\end{table*}

\begin{table*}[!h]
    \centering
    \scriptsize{
    \begin{tabular}{@{\:}l@{\:}cc@{\:\:\:}c@{\:\:\:}c@{\:\:\:}c@{\:\:\:}c@{\:\:\:}c@{\:\:\:}c@{\:\:\:}c@{\:\:\:}c@{\:}}
    \hline
        & CCA  & MVMD & MDDMp & MDDMf & wMLDAb & wMLDAe & wMLDAc  & wMLDAd & SSMLDR & NMLSDR \\
        \hline
Birds& 0.273& \textbf{0.419}& 0.407& 0.314& 0.250& 0.273& 0.250& 0.297& 0.203& \textbf{0.419} \\ 
Corel& 0.250& 0.261& 0.262& 0.252& 0.255& 0.254& 0.253& 0.267& 0.260& \textbf{0.283} \\ 
Emotions& 0.535& 0.673& 0.644& 0.564& 0.535& 0.589& 0.550& 0.589& 0.718& \textbf{0.728} \\ 
Enron& 0.620& 0.762& 0.762& 0.610& 0.587& 0.604& 0.606& 0.579& 0.544& \textbf{0.765} \\ 
Genbase& 0.950& 0.990& \textbf{0.995}& 0.955& 0.935& 0.960& 0.950& 0.965& 0.980& \textbf{0.995} \\ 
Medical& 0.583& 0.607& 0.592& 0.589& 0.538& 0.583& 0.577& 0.628& 0.323& \textbf{0.619} \\ 
Scene& 0.265& \textbf{0.732}& 0.729& 0.319& 0.258& 0.264& 0.247& 0.303& 0.656& 0.727 \\ 
Tmc2007& 0.527& 0.650& 0.648& 0.531& 0.523& 0.519& 0.516& 0.578& 0.604& \textbf{0.656} \\ 
Toy& 0.821& 0.888& 0.887& 0.819& 0.785& 0.821& 0.811& 0.850& 0.849& \textbf{0.903} \\ 
Yeast& \textbf{0.760}& 0.755& 0.749& 0.740& 0.747& 0.751& 0.748& 0.744& 0.751& 0.739 \\ 
\hline 
\# Best values& 1& 2& 1& 0& 0& 0& 0& 1& 0& \textbf{7} \\ 
Wilcoxon& 3.5& 8.0& 7.0& 4.0& 1.0& 3.5& 1.0& 5.0& 3.0& \textbf{9.0} \\
     \hline  
    \end{tabular}
    }
    \caption{Performance in terms of 1 - One error (OE') across 10 different benchmark datasets.}
    \label{tab:OE}
\end{table*}

\begin{table*}[!h]
    \centering
   \scriptsize{
    \begin{tabular}{@{\:}l@{\:}cc@{\:\:\:}c@{\:\:\:}c@{\:\:\:}c@{\:\:\:}c@{\:\:\:}c@{\:\:\:}c@{\:\:\:}c@{\:\:\:}c@{\:}}
    \hline
        & CCA  & MVMD & MDDMp & MDDMf & wMLDAb & wMLDAe & wMLDAc  & wMLDAd & SSMLDR & NMLSDR \\
        \hline
Birds& 0.821& 0.851& 0.852& 0.830& 0.818& 0.824& 0.824& 0.831& 0.808& \textbf{0.860} \\ 
Corel& 0.601& 0.617& 0.617& 0.603& 0.600& 0.599& 0.601& 0.603& 0.603& \textbf{0.628} \\ 
Emotions& 0.563& 0.684& 0.679& 0.579& 0.567& 0.565& 0.554& 0.587& 0.679& \textbf{0.696} \\ 
Enron& 0.738& 0.762& 0.763& 0.736& 0.737& 0.736& 0.734& 0.736& 0.724& \textbf{0.768} \\ 
Genbase& 0.983& 0.984& 0.984& 0.983& 0.985& 0.981& 0.981& 0.980& 0.985& \textbf{0.991} \\ 
Medical& 0.918& \textbf{0.941}& 0.939& 0.917& 0.909& 0.913& 0.911& 0.936& 0.859& 0.939 \\ 
Scene& 0.637& \textbf{0.899}& 0.898& 0.672& 0.625& 0.633& 0.624& 0.663& 0.860& 0.898 \\ 
Tmc2007& 0.740& 0.835& 0.835& 0.741& 0.740& 0.739& 0.741& 0.762& 0.790& \textbf{0.840} \\ 
Toy& 0.809& 0.837& 0.837& 0.807& 0.794& 0.805& 0.802& 0.822& 0.820& \textbf{0.849} \\ 
Yeast& 0.513& \textbf{0.533}& 0.532& 0.526& 0.526& 0.523& 0.519& 0.518& 0.530& 0.528 \\ 
\hline
\# Best values& 0& 3& 0& 0& 0& 0& 0& 0& 0& \textbf{7}\\ 
Wilcoxon& 2.5& 7.5& 7.5& 4.5& 2.0& 2.5& 1.5& 5.0& 3.0&\textbf{9.0} \\ 
     \hline  
    \end{tabular}
    }
    \caption{Performance in terms of 1 - Normalized coverage (Cov') across 10 different benchmark datasets.}
    \label{tab:COV}
\end{table*}

\begin{table*}[!h]
    \centering
   \scriptsize{
    \begin{tabular}{@{\:}l@{\:}cc@{\:\:\:}c@{\:\:\:}c@{\:\:\:}c@{\:\:\:}c@{\:\:\:}c@{\:\:\:}c@{\:\:\:}c@{\:\:\:}c@{\:}}
    \hline
        & CCA  & MVMD & MDDMp & MDDMf & wMLDAb & wMLDAe & wMLDAc  & wMLDAd & SSMLDR & NMLSDR \\
        \hline
Birds& \textbf{0.011}& 0.079& 0.076& 0.027& 0.002& 0.000& 0.000& 0.039& 0.006& 0.104 \\ 
Corel& 0.012& \textbf{0.023}& 0.022& 0.014& 0.010& 0.010& 0.010& 0.019& 0.010& 0.021 \\ 
Emotions& 0.381& 0.599& 0.604& 0.419& 0.366& 0.385& 0.371& 0.415& 0.623& \textbf{0.649} \\ 
Enron& 0.044& 0.102& \textbf{0.105}& 0.048& 0.043& 0.049& 0.044& 0.065& 0.063& 0.101 \\ 
Genbase& 0.520& 0.561& 0.603& 0.514& 0.497& 0.515& 0.497& 0.442& 0.558& \textbf{0.630} \\ 
Medical& 0.153& 0.168& 0.164& 0.159& 0.135& 0.126& 0.133& 0.197& 0.038& \textbf{0.175} \\ 
Scene& 0.059& \textbf{0.705}& 0.707& 0.132& 0.084& 0.055& 0.041& 0.098& 0.569& 0.700 \\ 
Tmc2007& 0.183& 0.419& 0.418& 0.189& 0.171& 0.177& 0.175& 0.212& 0.349& \textbf{0.434} \\ 
Toy& 0.732& 0.830& 0.828& 0.741& 0.709& 0.722& 0.724& 0.758& 0.776& \textbf{0.845} \\ 
Yeast& 0.266& 0.318& 0.323& 0.276& 0.281& 0.279& 0.248& 0.233& 0.321& \textbf{0.342} \\ 
\hline 
\# Best values& 0& 1& 2& 0& 0& 0& 0& 1& 0& \textbf{6} \\ 
Wilcoxon& 2.5& 7.5& 7.5& 5.0& 2.0& 2.0& 1.0& 3.5& 5.0& \textbf{9.0} \\ 
     \hline  
    \end{tabular}
    }
    \caption{Performance in terms of Macro F1-score (MaF1) across 10 different benchmark datasets.}
    \label{tab:MaF1}
\end{table*}

\begin{table*}[!h]
    \centering
    \scriptsize{
    \begin{tabular}{@{\:}l@{\:}cc@{\:\:\:}c@{\:\:\:}c@{\:\:\:}c@{\:\:\:}c@{\:\:\:}c@{\:\:\:}c@{\:\:\:}c@{\:\:\:}c@{\:}}
    \hline
        & CCA  & MVMD & MDDMp & MDDMf & wMLDAb & wMLDAe & wMLDAc  & wMLDAd & SSMLDR & NMLSDR \\
        \hline
Birds& 0.036& 0.178& 0.172& 0.063& 0.006& 0.000& 0.000& 0.065& 0.019& \textbf{0.197} \\ 
Corel& 0.017& \textbf{0.033}& 0.031& 0.019& 0.013& 0.013& 0.013& 0.031& 0.015& \textbf{0.033} \\ 
Emotions& 0.459& 0.630& 0.639& 0.450& 0.404& 0.448& 0.430& 0.460& 0.652& \textbf{0.666} \\ 
Enron& 0.351& 0.523& \textbf{0.530}& 0.413& 0.340& 0.378& 0.369& 0.310& 0.346& 0.518 \\ 
Genbase& 0.882& 0.953& 0.959& 0.872& 0.885& 0.902& 0.873& 0.881& 0.932& \textbf{0.968} \\ 
Medical& 0.459& 0.501& 0.495&\textbf{0.505}& 0.400& 0.440& 0.455& 0.498& 0.212& 0.496 \\ 
Scene& 0.066& 0.700& \textbf{0.702}& 0.142& 0.086& 0.058& 0.041& 0.102& 0.584& 0.698 \\ 
Tmc2007& 0.421& 0.589& 0.586& 0.443& 0.440& 0.438& 0.438& 0.485& 0.540& \textbf{0.590} \\ 
Toy& 0.729& 0.828& 0.826& 0.739& 0.706& 0.719& 0.721& 0.756& 0.774& \textbf{0.843} \\ 
Yeast& 0.573& 0.605& 0.607& 0.577& 0.582& 0.584& 0.555& 0.548& 0.609& \textbf{0.626} \\ 
\hline
\# Best values& 0& 1& 2& 1& 0& 0& 0& 0& 0& \textbf{7} \\ 
Wilcoxon& 2.5& 8.0& 7.5& 5.0& 1.5& 2.5& 2.0& 4.0& 3.5& \textbf{8.5} \\ 
     \hline  
    \end{tabular}
    }
    \caption{Performance in terms of Micro F1-score (MiF1) across 10 different benchmark datasets.}
    \label{tab:MiF1}
\end{table*}

Tab.~\ref{tab:HL} shows results in terms of HL'. NMLSDR gets best HL'-score for eight of the datasets and achieves a maximal Wilcoxon score, i.e performs statistically better than all nine other methods according to the test at a 5 \% significance level. The second best method MDDMp gets the highest HL' score for three datasets and Wilcoxon score of 7.5.
From Tab.~\ref{tab:RL} we see that NMLSDR achieves the highest RL'-score seven times and a Wilcoxon score of 8.5. The second best method is MVMD, which obtains three of the highest RL' values and a Wilcoxon score of 8.0.

Tab.~\ref{tab:AP} shows performance in terms of AP. The highest AP score is achieved for NMLSDR for eight datasets and it gets a maximal Wilcoxon score of 9.0. According to the Wilcoxon score second place is tied between MVMD and MDDMp. However, MVMD gets the highest AP score for two datasets, whereas MDDMp does not get the highest score for any of them.
OE' is presented in Tab.~\ref{tab:OE}. We can see that NMLSDR gets a maximal Wilcoxon score and the highest OE' score for seven datasets. MVMD is number two with a Wilcoxon score of 8.0 and two best values.

Tab.~\ref{tab:COV} shows Cov'. NMLSDR gets a maximal Wilcoxon score and the highest Cov' value for seven datasets. Despite that MVMD gets the highest Cov' for three datasets and MDDMp for none of the datasets, the second  best Wilcoxon score is 7.5 and tied between MVMD and MDDMp.
MaF1 is shown in Tab.~\ref{tab:MaF1}. The best method, which is our proposed method gets a maximal Wilcoxon score and the highest MaF1 value for six datasets.
Tab.~\ref{tab:MiF1} shows MiF1. NMLSDR achieves 8.5 in Wilcoxon score and has the highest MiF1 score for seven datasets.

\begin{figure}[t]
    \centering
            \includegraphics[trim = 44mm 85mm 44mm 88mm, clip, width=.5\linewidth]{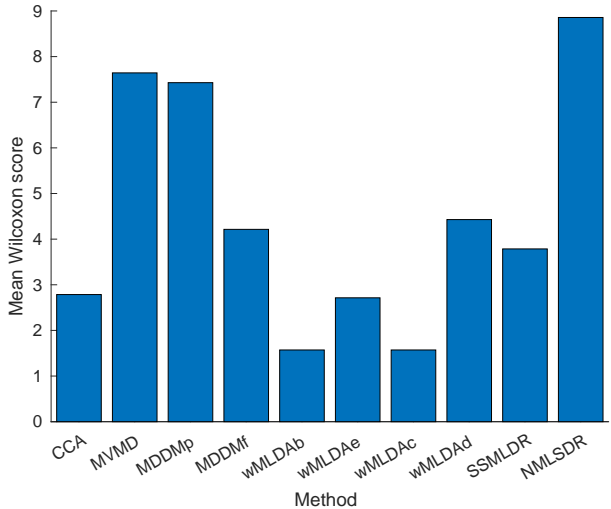} 
    \caption{Mean of the Wilcoxon score obtained over the 7 different metrics.}
\label{fig:Wilcoxon}
\end{figure}

In total, NMLSDR consistently gives the best performance for all seven evaluation metrics. 
Moreover, in order to summarize our findings, we compute the mean Wilcoxon score across all seven performance metrics and plot the result in Fig.~\ref{fig:Wilcoxon}.  If we sort these results, we get NMLSDR (8.86), MVMD (7.64), MDDMp (7.43), wMLDAd (4.43), MDDMf (4.21), SSMLDR (3.79), CCA (2.79), wMLDAe (2.71) and wMLDAb/wMLDAc (1.57). The best method, which is our proposed method, gets a mean value that is 1.22 higher than number two. The second best method is MVMD, slightly better than MDDMp. The best MLDA-based method is wMLDAd, which is ranked 4th, however, with a much lower mean value than the three best methods. The semi-supervised extension of MLDA (SSMLDR) is ranked 6th and is actually performing worse that wMLDAd, which is a bit surprising. However, SSMLDR also uses a binary weighting scheme, and should therefore be considered as a semi-supervised variant of wMLDAb, which it performs considerably better than.  wMLDAb and wMLDAc give the worst performance of all the 10 methods.

The main reason why the MLDA-based approaches in general perform worse than the other DR methods is probably related to what we discussed in Sec.~\ref{sec: related work}, namely that LDA-based approaches are heavily affected by outliers and wrongly labeled data. More concretely, the fact that the number of labeled data points are relatively few and that the labels are noisy, leads to errors in the scatter matrices that even might amplify since one has to invert a matrix to solve the generalized eigenvalue problem.  
The semi-supervised extension of MLDA, SSMLDR, improves quite much compared to wMLDAb, but the starting point is so bad that even though it improves, it cannot compete with the best methods. On the other hand, the MDDM-based methods (MVMD and MDDMp) are not so sensitive to  label noise and the fact that there are few labels, and therefore these methods can perform quite well even though they are trained only on the labeled subset. Hence, the reasons to the good performance of NMLSDR are probably that  MDDMp is the basis of NMLSDR, and that NMLSDR in addition uses our label propagation method to improve.

\section{Case study}
\label{sec: Case study}

\begin{figure}
    \centering
    \includegraphics[trim = 0mm 0mm 0mm 0mm, clip,width = \linewidth]{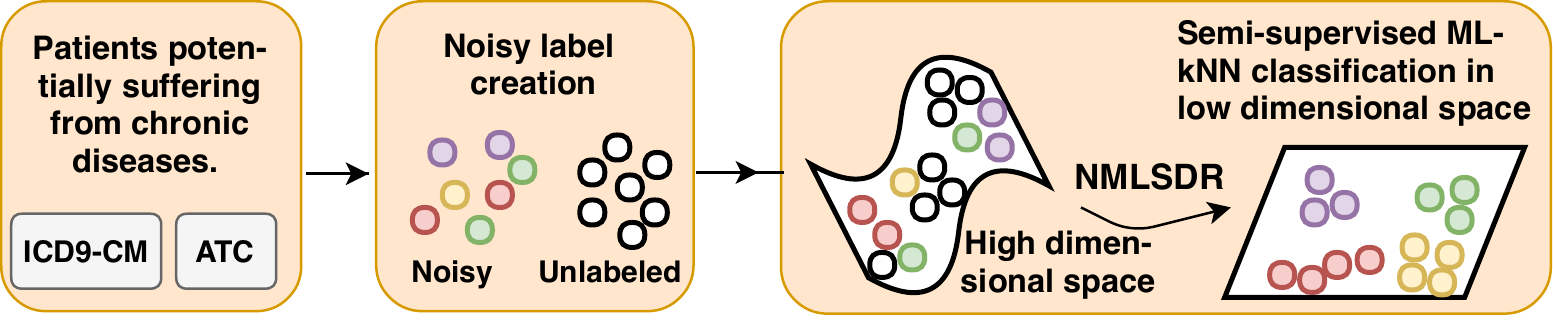}
    \caption{Illustration of proposed framework applied to identify patients with chronic diseases.}
    \label{fig:schematics}
\end{figure}

In this section, we describe a case study where we study patients  potentially suffering from multiple chronic diseases. This healthcare case study reflects the need for label noise-tolerant methods in a non-standard situation (semi-supervised learning, multiple labels, high dimensionality).  The objective is to identify patients with certain chronic diseases, more specifically hypertension and/or diabetes mellitus. In order to do so, we take an approach where we use clinical expertise to create a partially and noisy labeled dataset, and thereafter apply our proposed end-to-end framework, namely NMLSDR for dimensionality reduction in combination with semi-supervised ML-kNN to classify these patients. An overview of the framework employed in the case study is shown in Fig.~\ref{fig:schematics}.

\paragraph{Chronic diseases}
According to The World Health Organisation, a disease is defined as chronic if  one or several of the following criteria are satisfied:  the disease is  permanent, requires special training of the patient
for rehabilitation, is caused by non-reversible pathological alterations, or  requires a long period of supervision, observation, or care. The two most prevalent chronic diseases for people over 64 years are those that we study in this paper, namely hypertension and diabetes mellitus~\cite{soni2001}.
These types of diseases represent an increasing problem in modern societies all over the world, which to a large degree is due to a general increase in life expectancy, along with an increased prevalence of chronic diseases in an aging population~\cite{calderon2016assessing}.
Moreover, the economical burden associated with these chronic conditions is high. For example, in 2017, treatment of diabetic patients accounted for 1 out of 4 healthcare dollars in the United States~\cite{american2018economic}. 
Hence, in the future, a significant amount of resources must be devoted to the care of chronic patients and it will be important not only to improve  the patient care, but also more efficiently  allocate the resources spent on treatment of these diseases.

\subsection{Data}
In this case study, we study a dataset consisting of patients that potentially have one or more chronic diseases. All of these patients got some type of treatment at University Hospital of Fuenlabrada, Madrid (Spain) in the year 2012.  The patients are described by diagnosis codes following the International Classification of Diseases 9th revision, Clinical Modification (ICD9-CM)~\cite{ICD9-CM}, and pharmacological dispensing codes according to Anatomical Therapeutic Chemical (ATC) classification systems~\cite{world2016guidelines}.  Some preprocessing steps are considered. Similarly to~\cite{soguero2018use, sanchez2018scaled},  the ICD9-CM and ATC codes are represented using frequencies, i.e, for each patient, we consider all encounters with the health system in 2012 and we count how many times each ICD9-CM and ATC code appear in the electronic health record.  In total there are 1517 ICD9-CM codes and 746 ATC codes.
However, all codes that appear for less than 10 patients across the training set are removed. After this feature selection, the dimensionality of the data is 455, of which 267 represent ICD9-CM codes and 188 represent ATC codes.

We do have access to ground truth labels that indicate what type of chronic disease(s) the patients have. These are provided by a patient classification system developed by the company 3M~\cite{Averill1999}. This classification system stratify patients into so-called Clinical Risk Groups (CRG) that indicate what type(s) of chronic disease the patient has and the severity based on the patient encounters with the health system during a period of time, typically one year. A five-digit classification code is used to assign each patient to a severity risk group. The first digit of the CRG is the core health status group, ranging from healthy (1) to catastrophic (9); the second to fourth digits represents the base 3M CRG; and the fifth digit is used for characterizing the severity-of-illness levels. 

For the purpose of this work, the ground truth labels are only used for cohort selection and final evaluation of our models. For the remaining parts they are considered unknown.
To select a cohort,  we consider the first four digits of the CRGs to analyze the the following chronic conditions: CRG-1000 (healthy), which contains 46835 individuals; CRG-5192 (hypertension) with 12447 patients;  CRG-5424 (diabetes), which has 2166 patients; and  CRG-6144 (hypertension and diabetes), with a total of 3179 patients. 
We employ an undersampling strategy and randomly select 2166 patients from each of the four categories, and thereby obtain balanced classes. 
An independent test set is created by randomly selecting 20 \% of these patients. Hence, the training set contains 6932 patients and the test set 1732 patients.

\subsection{Rule-based creation of noisy labeled training data using clinical knowledge} 
\label{Sec: label generation}

There are some important ICD9-CM codes and ATC-drugs that are strongly correlated with hypertension and diabetes, respectively. These are verified by our clinical experts and described in Tab.~\ref{tab:IMP_icd}. In particular, the ICD9-CM code 250 is important for diabetes because it is the code for \emph{diabetes mellitus}. 
Similarly, the ICD9-CM codes 401-405 are important for hypertension because they describe different types of hypertension.

\begin{table}[!t]
    \centering
     \scriptsize{
    \begin{tabular}{|l|l|l|}
    \hline
       Chronicity  & ATC codes & ICD9-CM codes  \\
       \hline
    Hypertension  & C01AA,  C01BA, C01BA, C01BC, C01BD, C01CA, C01CB, C01CX,  & 362, \\ 
     & C01DA, C01DX, C01EB, C02AB, C02AC, C02CA, C02DB, C02DC, & 401, \\
     & C02DD, C02K, C02LC, C03AA, C03AX, C03BA, C03CA, C03DA & 402, \\
     & C03EA, C03EB, C04AD, C04AE, C04AX, C05AA, C05AD, C05AE, & 403, \\
     & C05AX, C05BA, C05BB, C05BX, C05CA, C05CX, C07AA, C07AB, & 404, \\ 
     & C07AG, C07B, C07G, C07D, C07E, C07X, C08CA, C08DA, C08DB, & 405, \\ 
     & C08GA, C09AA, C09BA, C09BB, C09CA, C09DA, C09DB, C09XA, & 760  \\
     & C10AA, C10AB, C10AC, C10AD, C10AX, C10BA, C10BX & \\
     \hline 
    Diabetes & A10AB, A10AC, A10AD, A10AE, A10AF, A10BA, A10BB, & 250, 588, \\
    & A10BD, A10BFM, A10BGM, A10BH, A10BX, &  648, 775  \\
    \hline
    \end{tabular}
    }
    \caption{ICD9-CM codes and ATC codes associated with hypertension and diabetes.}
    \label{tab:IMP_icd}
\end{table}

In this case study we are interested in four groups, namely those that have hypertension, those that have diabetes, those that have both, and those that do not have any these two chronic diseases. Thanks to the clinical expertise and the information that they provided us with, which is summarized in Tab.~\ref{tab:IMP_icd}, we can create a partially and noisy labeled dataset using the following set of rules.
\begin{enumerate}
    \item Those that have the ICD codes 250 and any of the codes 401-405 are assigned to both the hypertension and diabetes class. 
    \item Those that have the ICD code 250, but none of the 7 ICD9-CM codes and 64 ATC drugs listed by the clinicians as indicators for hypertension, are labeled with diabetes.
    \item Those that have any of the ICD9-CM codes 401-405, but none of the 4 ICD9-CM codes for diabetes or 12 ATC drugs for diabetes, are labeled with hypertension.
    \item Those that do not have any of the ICD9-CM codes or ATC drugs listed up in Tab.~\ref{tab:IMP_icd} are labeled as healthy. 
    \item The remaining patients do not get a label.
\end{enumerate}

In total, this leads to 1734 in the healthy class, 2547 in the hypertension class, 1971 in the diabetes class. 1302 of the patients in the hypertension class also belongs to the diabetes class. 1982 of the patients do not get a label using the the routine described above.
To be able to examine for statistical significance, we randomly select 1000 of the noisy labeled patients and 1000 of the unlabeled patients. By doing so, we can repeat the experiments several times and test for significance using a pairwise t-test. We do the repetition 10 times and let the significance level be 95\%.

\subsubsection{Performing feature extraction and classification}
After having obtained the partially and noisy labeled multi-label dataset, we do feature extraction using NMLSDR, followed by semi-supervised multi-label classification, exactly in the same manner as we did it for the synthetic toy data in Section~\ref{sec: illustrative example}.
In this case study, we use the same evaluation metrics, hyper-parameters and baseline feature extraction methods as explained in Sec.~\ref{sec: evaluation metrics}.  The dimensionality of the embedding is set to $2$ for all embedding methods.

\subsection{Results}

Tab.~\ref{tab:classication_chronicity} shows the performance of the different DR methods on the task of classifying patients with chronic diseases in terms of seven different evaluation metrics. According to the pairwise t-test, our method achieves the best performance for all metrics. Second place is tied between MDDMp and MVMD. The semi-supervised variant of MLDA, namely SSMLDR, performs better than the supervised counterparts (wMLDAb, wMLDAc, wMLDAd, wMLDAe) and is consistently ranked 4th according to all metrics. Interestingly, the more advanced weighting schemes in wMLDAc and wMLDAd actually lead to worse results than what the simple weights in wMLDAb and wMLdAe give.   CCA gives the worst performance according to 4 of the evaluation measures, for the 3 other measures the difference between CCA and wMLDAd is not significant.

\begin{table}[t]
    \centering
    \tiny{
    \begin{tabular}{@{ }lc@{\:\:\:\:\:\:}c@{\:\:\:\:\:\:}c@{\:\:\:\:\:\:}c@{\:\:\:\:\:\:}c@{\:\:\:\:\:\:}c@{\:\:\:\:\:\:}c@{ }}
    \hline
        \textbf{Method} & HL'  & RL' & AP & OE' & Cov' & MaF1 & MiF1  \\
     \hline
CCA& 0.782 $\pm$ 0.009& 0.823 $\pm$ 0.008& 0.866 $\pm$ 0.006& 0.755 $\pm$ 0.011& 0.798 $\pm$ 0.004& 0.712 $\pm$ 0.012& 0.741 $\pm$ 0.011\\ 
MVMD& 0.875 $\pm$ 0.006& 0.930 $\pm$ 0.006& 0.942 $\pm$ 0.004& 0.894 $\pm$ 0.006& 0.861 $\pm$ 0.005& 0.853 $\pm$ 0.008& 0.858 $\pm$ 0.006\\ 
MDDMp& 0.875 $\pm$ 0.006& 0.930 $\pm$ 0.005& 0.942 $\pm$ 0.003& 0.895 $\pm$ 0.006& 0.861 $\pm$ 0.005& 0.853 $\pm$ 0.008& 0.858 $\pm$ 0.006\\ 
MDDMf& 0.811 $\pm$ 0.010& 0.853 $\pm$ 0.012& 0.888 $\pm$ 0.009& 0.798 $\pm$ 0.017& 0.815 $\pm$ 0.006& 0.750 $\pm$ 0.015& 0.774 $\pm$ 0.013\\ 
wMLDAb& 0.794 $\pm$ 0.007& 0.844 $\pm$ 0.012& 0.883 $\pm$ 0.008& 0.788 $\pm$ 0.017& 0.810 $\pm$ 0.008& 0.731 $\pm$ 0.012& 0.744 $\pm$ 0.011\\ 
wMLDAe& 0.805 $\pm$ 0.008& 0.856 $\pm$ 0.009& 0.891 $\pm$ 0.006& 0.801 $\pm$ 0.014& 0.818 $\pm$ 0.005& 0.749 $\pm$ 0.013& 0.763 $\pm$ 0.012\\ 
wMLDAc& 0.790 $\pm$ 0.007& 0.842 $\pm$ 0.008& 0.882 $\pm$ 0.004& 0.783 $\pm$ 0.009& 0.810 $\pm$ 0.005& 0.729 $\pm$ 0.012& 0.745 $\pm$ 0.011\\ 
wMLDAd& 0.779 $\pm$ 0.013& 0.838 $\pm$ 0.012& 0.874 $\pm$ 0.008& 0.770 $\pm$ 0.016& 0.805 $\pm$ 0.008& 0.720 $\pm$ 0.017& 0.729 $\pm$ 0.018\\ 
SSMLDR& 0.839 $\pm$ 0.005& 0.889 $\pm$ 0.009& 0.911 $\pm$ 0.006& 0.839 $\pm$ 0.012& 0.835 $\pm$ 0.008& 0.799 $\pm$ 0.007& 0.811 $\pm$ 0.005\\ 
NMLSDR& \textbf{0.882 $\pm$ 0.005} & \textbf{0.939 $\pm$ 0.004}& \textbf{0.950 $\pm$ 0.003}& \textbf{0.909 $\pm$ 0.006}& \textbf{0.867 $\pm$ 0.005}& \textbf{0.864 $\pm$ 0.007}& \textbf{0.865 $\pm$ 0.005} \\ 
     \hline  
    \end{tabular}
    }
    \caption{Results in terms of 7 evaluation measures (average$\pm$std) obtained by doing feature extraction using different methods, followed by semi-supervised ML-kNN classification, on partially and noisy labeled chronicity data. The best performing methods according to each of the 7 metrics are marked in bold, where the statistical significance is examined using a pairwise t-test at 95\% significance level.}
    \label{tab:classication_chronicity}
\end{table}

\begin{figure}[t]
    \centering
        \begin{subfigure}{.49\textwidth}
            \includegraphics[trim = 41mm 87mm 44mm 86mm, clip,height = 5cm]{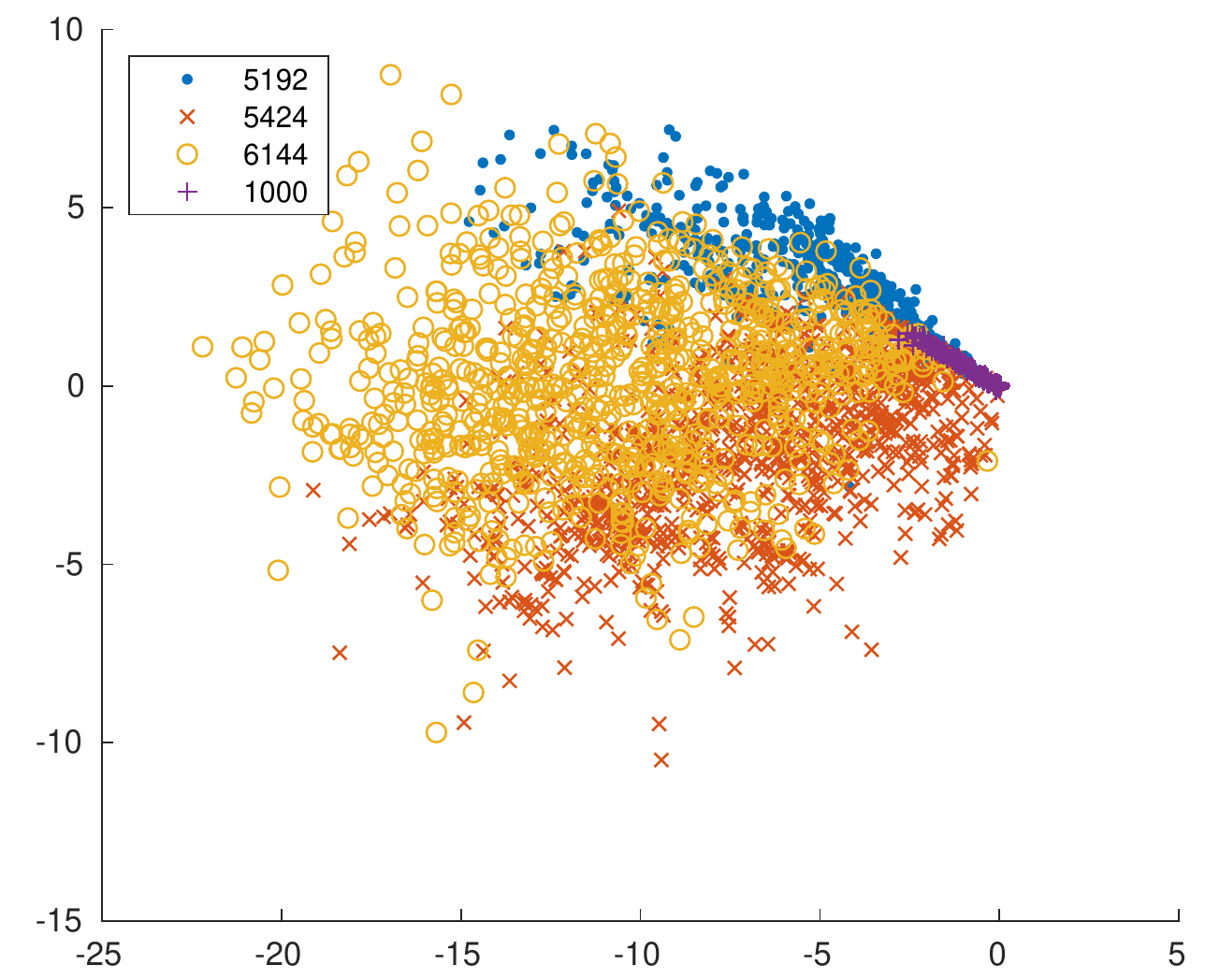} 
            \caption{}
            \label{fig:chronicity_MDDMp}
        \end{subfigure}    
        \begin{subfigure}{.49\textwidth}
            \includegraphics[trim = 41mm 87mm 44mm 86mm, clip,height = 5cm]{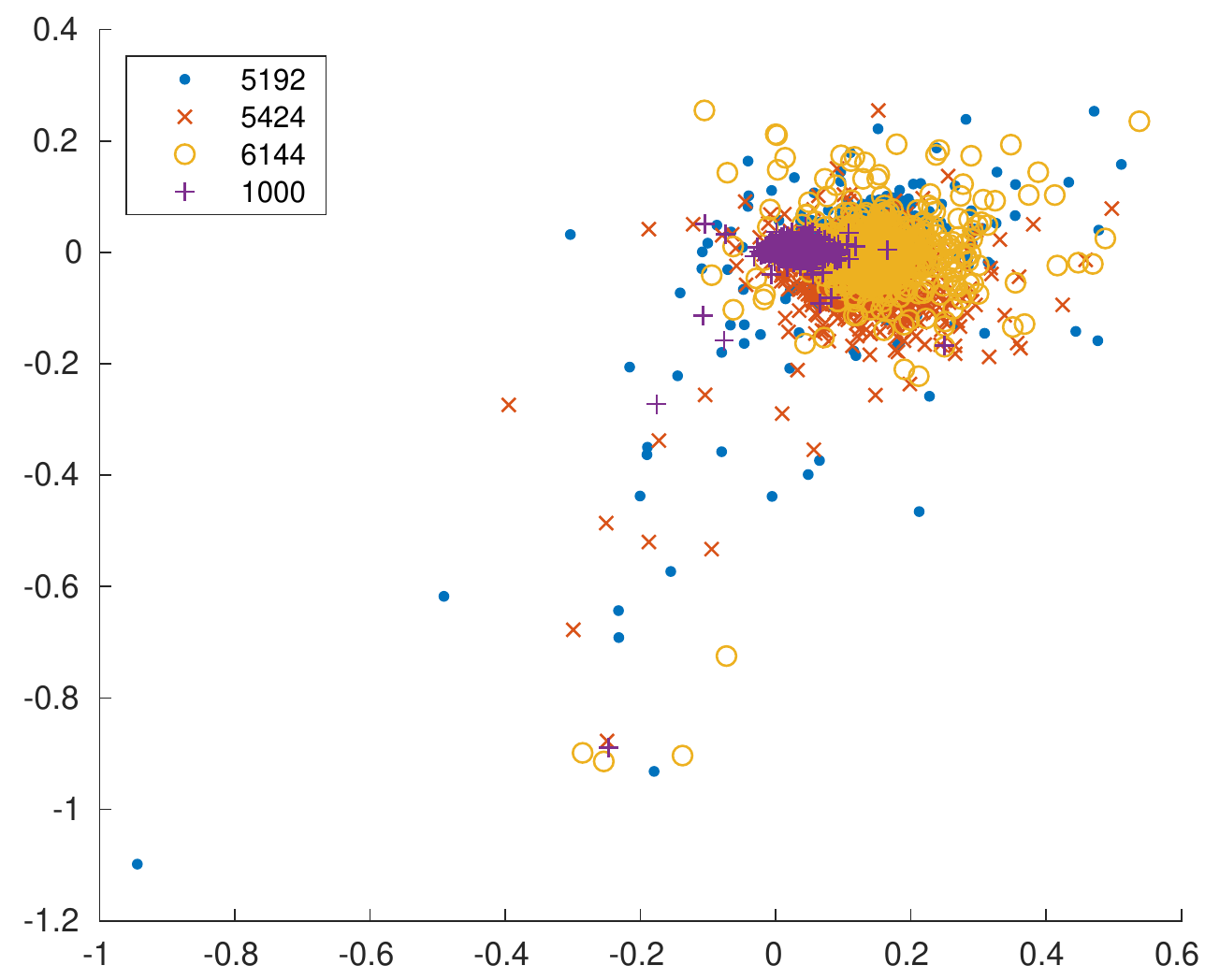} 
            \caption{}
            \label{fig:chronicity_MLDA}
        \end{subfigure}  
        \begin{subfigure}{.49\textwidth}
            \includegraphics[trim = 41mm 87mm 44mm 86mm, clip,height = 5cm]{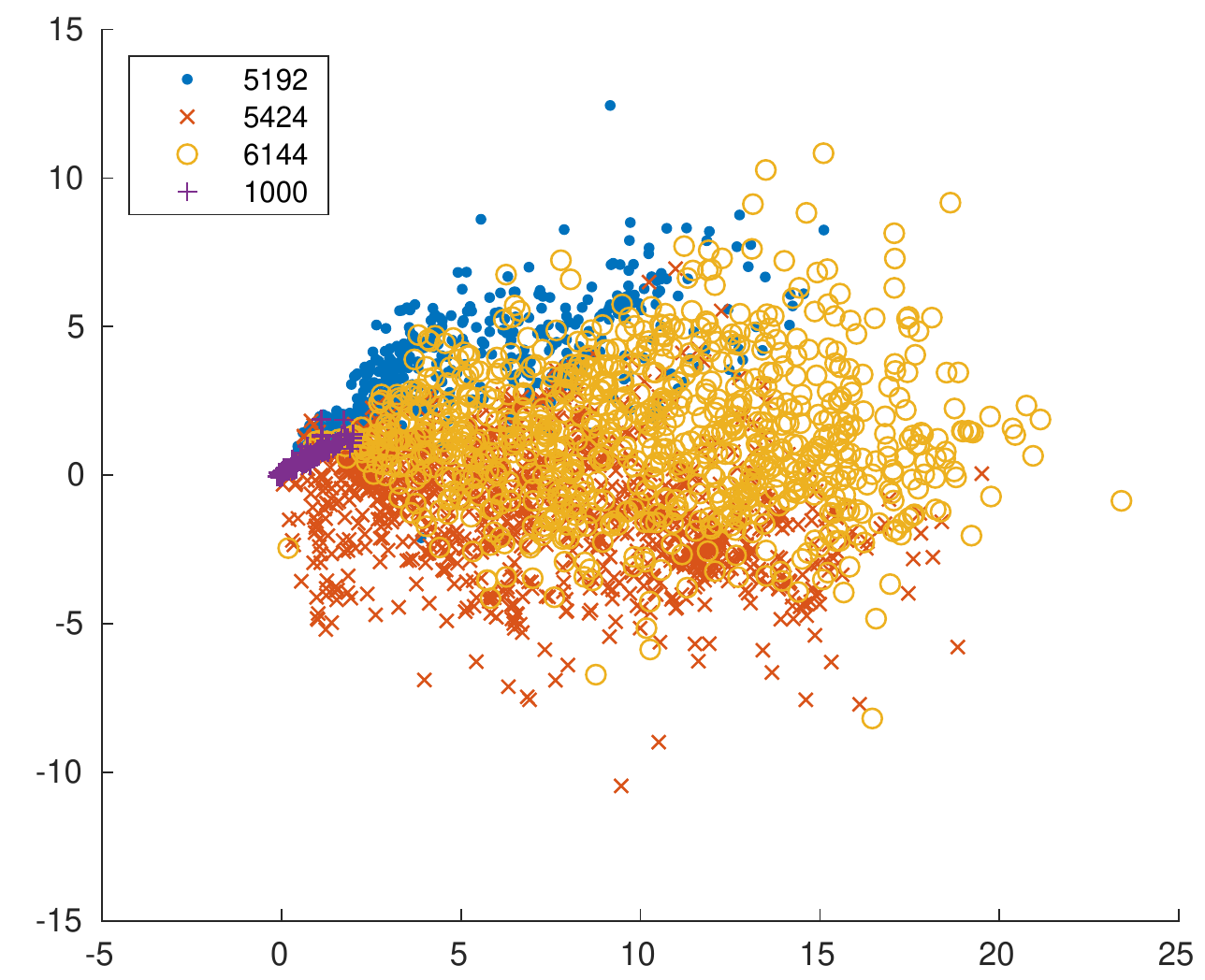} 
            \caption{}
            \label{fig:chronicity_NMLSDR}
        \end{subfigure}    
        \begin{subfigure}{.49\textwidth}
            \includegraphics[trim = 41mm 87mm 44mm 86mm, clip,height = 5cm]{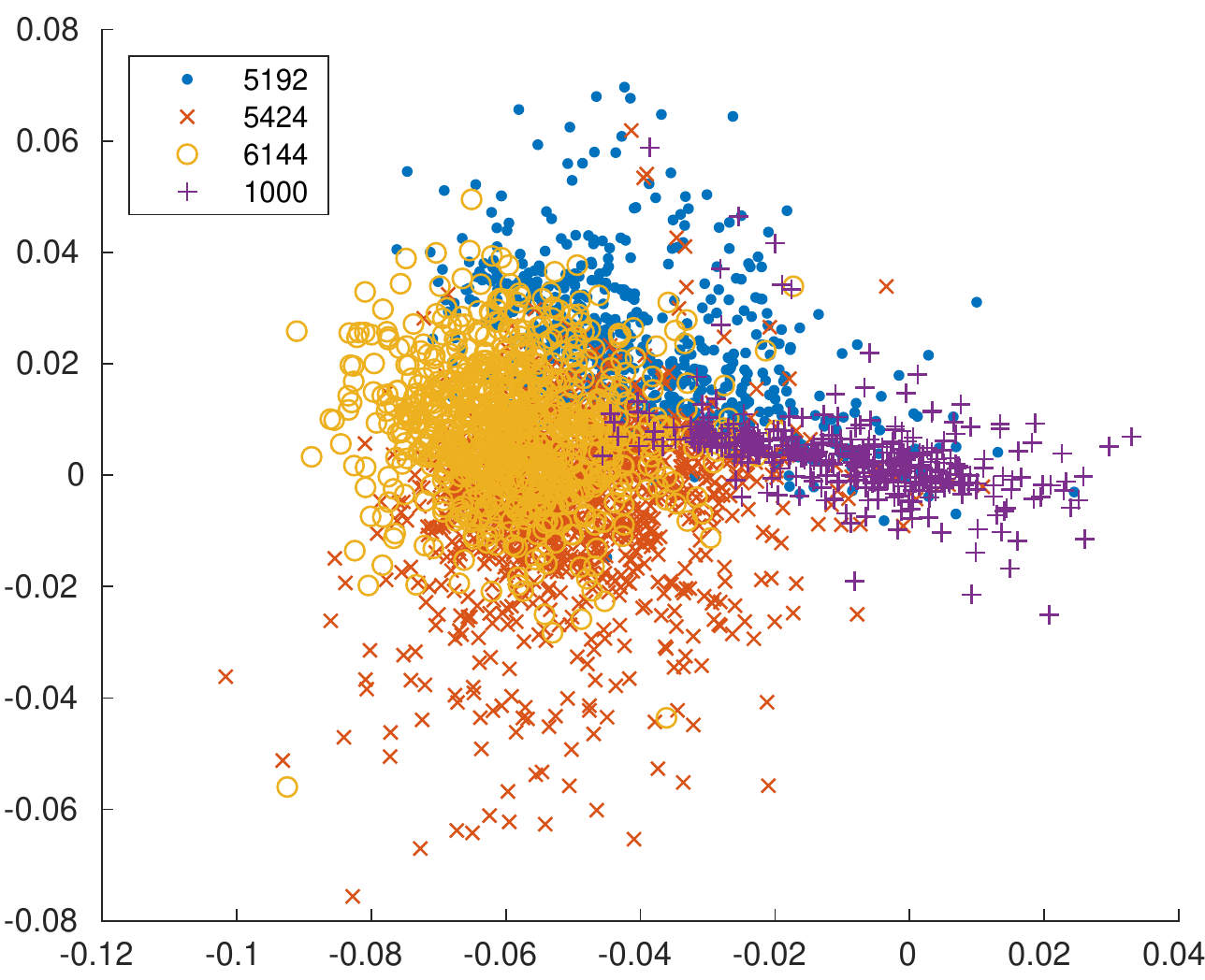} 
            \caption{}
            \label{fig:chronicity_SSMLDR}
        \end{subfigure}         
    \caption{Plot two-dimensional embeddings of the chronic patients obtained using four different DR methods:  (a) MDDMp. (b) wMLDAb (c) NMLSDR (d) SSMLDR. 
    The different colors and markers represent the true CRG-labels of the patients.}
\label{fig:chronicity}
\end{figure}

Fig.~\ref{fig:chronicity} shows plots of the two-dimensional embeddings of the chronic patients obtained using four different DR methods, namely MDDMp, wMLDAb, NMLSDR and SSMLDR. The different colors and markers represent the true CRG-labels of the patients. As we can see, visually the MDDMp and NMLSDR embeddings look quite similar. The healthy patients are squeezed together in a small area (purple dots), and the yellow dots that represent patients that have both diabetes and hypertension are placed between the blue dots, which are those that have only hypertension, and the red dots, which represent the patient that only have diabetes. Intuitively, this placement makes sense. On the other hand, the embedding obtained using SSMLDR does not look similar to its counterpart obtained using wMLDAb, and it is easy to see why the performance of wMLDAb is worse.

\section{Conclusions and future work}
\label{sec: Conclusion} 

In this paper we have introduced the NMLSDR method, a dimensionality reduction method for partially and noisy labeled multi-label data. To our knowledge, NMLSDR is the only method the can explicitly deal with this type of data. Key components in the method are a label propagation algorithm that can deal with noisy data and maximization of feature-label dependence using the Hilbert-Schmidt independence criterion. Our extensive experimental sections show that NMLSDR is a good dimensionality reduction method in settings where one has access to partially and noisy labeled multi-label data.

A potential limitation of NMLSDR is that it is a linear dimensionality reduction method. The method can, however, be extended within the framework of kernel methods~\cite{10.1007/BFb0020217, MIKALSEN2018569, HOOSHMANDMOGHADDAM2016921} to deal with non-linear data. In fact, NMLSDR is already a kernel method in the current formulation, in which we put a linear kernel over the feature space. The linear kernel can, however, straightforwardly be replaced with a non-linear kernel. The effect of doing this will be investigated in future work.
In the future, we will also investigate more thoroughly the effect of using different weighting schemes in NMLSDR, similarly to how it is done in MLDA with wMLDAb, wMLDAc, wMLDAd and wMDLAd.

It should be noticed that in our experiments, in addition to evaluating the proposed method visually for a couple of the datasets, we combined the NMLSDR with a popular multi-label classifier, namely the multi-label k-nearest neighbor classifier. By doing so, we could quantitatively evaluate the quality of the embeddings learned by the  NMLSDR and compare to alternative dimensionality reduction methods.
However,  many other multi-label classifiers exist~\cite{tahir2012multilabel, madjarov2012extensive, XU2013885,CHEN201661,LIU2018307,WANG20143405, trajdos2015extension,TRAJDOS201860,ZHUANG2018225}. As future work, it would be interesting to investigate if the proposed method outperforms alternative dimensionality reduction methods in conjunction with other classifiers as well.

Further, we recognize that the outcome of label propagation using a graph is influenced by several factors. More precisely, there are two main components that affect how the labels propagate, namely the particular method chosen and how the graph is constructed. Both of these two components are important, as discussed in~\cite{zhu2005semi, zhu2006semi}. In our experiments, we chose a neighborhood graph with binary weights. However, in future work it would be interesting to more thoroughly investigate the sensitivity of NMLSDR  with respect to the particular choices made for constructing the graph.

\section*{Acknowledgments}
This work was partially funded by the Norwegian Research Council FRIPRO grant no. 239844 on developing the \emph{Next Generation Learning Machines}. Cristina Soguero-Ruiz is partially supported by project TEC2016-75361-R from Spanish Government and by Project DTS17/00158 from Institute of Health Carlos III (Spain). The authors would like to thank clinicians at University Hospital of Fuenlabrada for their helpful advice and comments on various issues examined in the case study.

\section*{References}
\bibliographystyle{elsarticle-num}
\bibliography{references}

\end{document}